\documentclass[sigconf]{acmart}

\copyrightyear{2024}
\acmYear{2024}
\setcopyright{acmlicensed}\acmConference[MM '24]{Proceedings of the 32nd ACM International Conference on Multimedia}{October 28-November 1, 2024}{Melbourne, VIC, Australia}
\acmBooktitle{Proceedings of the 32nd ACM International Conference on Multimedia (MM '24), October 28-November 1, 2024, Melbourne, VIC, Australia}
\acmDOI{10.1145/3664647.3680809}
\acmISBN{979-8-4007-0686-8/24/10}

\AtBeginDocument{%
  }

\usepackage{balance}
\usepackage{amsmath}
\usepackage{tablefootnote}
\usepackage{colortbl}
\usepackage{indentfirst} 
\usepackage{amssymb}
\usepackage{bbding}
\usepackage{graphbox}
\usepackage{graphicx}
\usepackage{caption}
\usepackage{subcaption}

\usepackage{multirow}
\usepackage{stfloats}
\usepackage{url}
\usepackage{tabularx}
\usepackage[linesnumbered,ruled,vlined]{algorithm2e}

\newcommand{\eat}[1]{}
\acmConference[MM '24]{Proceedings of the 32nd ACM International Conference on Multimedia}{October 28-November 1, 2024}{Melbourne, VIC, Australia}
\acmISBN{979-8-4007-0686-8/24/10}

\begin{document}

\newcounter{alphasect}
\def\alphainsection{0}

\let\oldsection=\section
\def\section{%
  \ifnum\alphainsection=1%
    \addtocounter{alphasect}{1}
  \fi%
\oldsection}%

\renewcommand\thesection{%
  \ifnum\alphainsection=1% 
    \Alph{alphasect}
  \else%
    \arabic{section}
  \fi%
}%

\newenvironment{alphasection}{%
  \ifnum\alphainsection=1%
    \errhelp={Let other blocks end at the beginning of the next block.}
    \errmessage{Nested Alpha section not allowed}
  \fi%
  \setcounter{alphasect}{0}
  \def\alphainsection{1}
}{%
  \setcounter{alphasect}{0}
  \def\alphainsection{0}
}%

\title{Caterpillar: A Pure-MLP Architecture with Shifted-Pillars-Concatenation}


% \affiliation{%

\author{Jin Sun}
\affiliation{%
  \institution{School of Computer Science and
 Engineering, University of Electronic Science and Technology of China}
  \city{Chengdu}
  \country{China}}
\email{sunjin@uestc.edu.cn}

\author{Xiaoshuang Shi}
\authornote{Corresponding author.}
\affiliation{%
  \institution{School of Computer Science and
 Engineering, University of Electronic Science and Technology of China}
  \city{Chengdu}
  \country{China}}
\email{xsshi2013@gmail.com}

\author{Zhiyuan Wang}
\affiliation{%
  \institution{School of Computer Science and
 Engineering, University of Electronic Science and Technology of China}
  \city{Chengdu}
  \country{China}}
\email{yhzywang@gmail.com}

% Philadelphia, PA, USA
\author{Kaidi Xu}
\affiliation{%
  \institution{Department of Computer Science, Drexel University} 
  \city{Philadelphia}
  \country{PA, USA}}
\email{xu.kaid@husky.neu.edu}

\author{Heng Tao Shen}
\affiliation{
  \institution{School of Computer Science and
 Engineering, University of Electronic Science and Technology of China}
  \city{Chengdu}
  \country{China}}
\affiliation{%
  \institution{Tongji University}
  \city{Shanghai}
  \country{China}
}
\email{shenhengtao@hotmail.com}

\author{Xiaofeng Zhu}
\affiliation{%
  \institution{School of Computer Science and Engineering, University of Electronic Science and Technology of China}
  \city{Chengdu}
  \country{China}
}
\email{seanzhuxf@gmail.com}

\renewcommand{\shortauthors}{Jin Sun et al.}

\begin{abstract}
Modeling in Computer Vision has evolved to MLPs.
Vision MLPs naturally lack local modeling capability, to which the simplest treatment is combined with convolutional layers.
Convolution, famous for its sliding window scheme, also suffers from this scheme of redundancy and lower parallel computation.
In this paper, we seek to dispense with the windowing scheme and introduce a more elaborate and parallelizable method to exploit locality. 
To this end, we propose a new MLP module, namely Shifted-Pillars-Concatenation (SPC), that consists of two steps of processes: (1) Pillars-Shift, which generates four neighboring maps by shifting the input image along four directions, and (2) Pillars-Concatenation, which applies linear transformations and concatenation on the maps to aggregate local features. SPC module offers superior local modeling power and performance gains, making it a promising alternative to the convolutional layer.
Then, we build a pure-MLP architecture called Caterpillar by replacing the convolutional layer with the SPC module in a hybrid model of sMLPNet \cite{tang2022sMLPNet}. 
Extensive experiments show Caterpillar's excellent performance on both small-scale and ImageNet-1k classification benchmarks, with remarkable scalability and transfer capability possessed as well.
The code is available at \emph{\url{https://github.com/sunjin19126/Caterpillar}}.
\end{abstract}

\begin{CCSXML}
<ccs2012>
 <concept>
  <concept_id>00000000.0000000.0000000</concept_id>
  <concept_desc>Do Not Use This Code, Generate the Correct Terms for Your Paper</concept_desc>
  <concept_significance>500</concept_significance>
 </concept>
 <concept>
  <concept_id>00000000.00000000.00000000</concept_id>
  <concept_desc>Do Not Use This Code, Generate the Correct Terms for Your Paper</concept_desc>
  <concept_significance>300</concept_significance>
 </concept>
 <concept>
  <concept_id>00000000.00000000.00000000</concept_id>
  <concept_desc>Do Not Use This Code, Generate the Correct Terms for Your Paper</concept_desc>
  <concept_significance>100</concept_significance>
 </concept>
 <concept>
  <concept_id>00000000.00000000.00000000</concept_id>
  <concept_desc>Do Not Use This Code, Generate the Correct Terms for Your Paper</concept_desc>
  <concept_significance>100</concept_significance>
 </concept>
</ccs2012>
\end{CCSXML}

\ccsdesc[500]{Multimedia Foundation Models~Vision and Language}

\keywords{Computer Vision, Pure-MLP Architecture, Caterpillar, SPC Module}


\maketitle

\section{Introduction}

\begin{figure}[t]
  \centering
   \includegraphics[width=0.95 \linewidth]{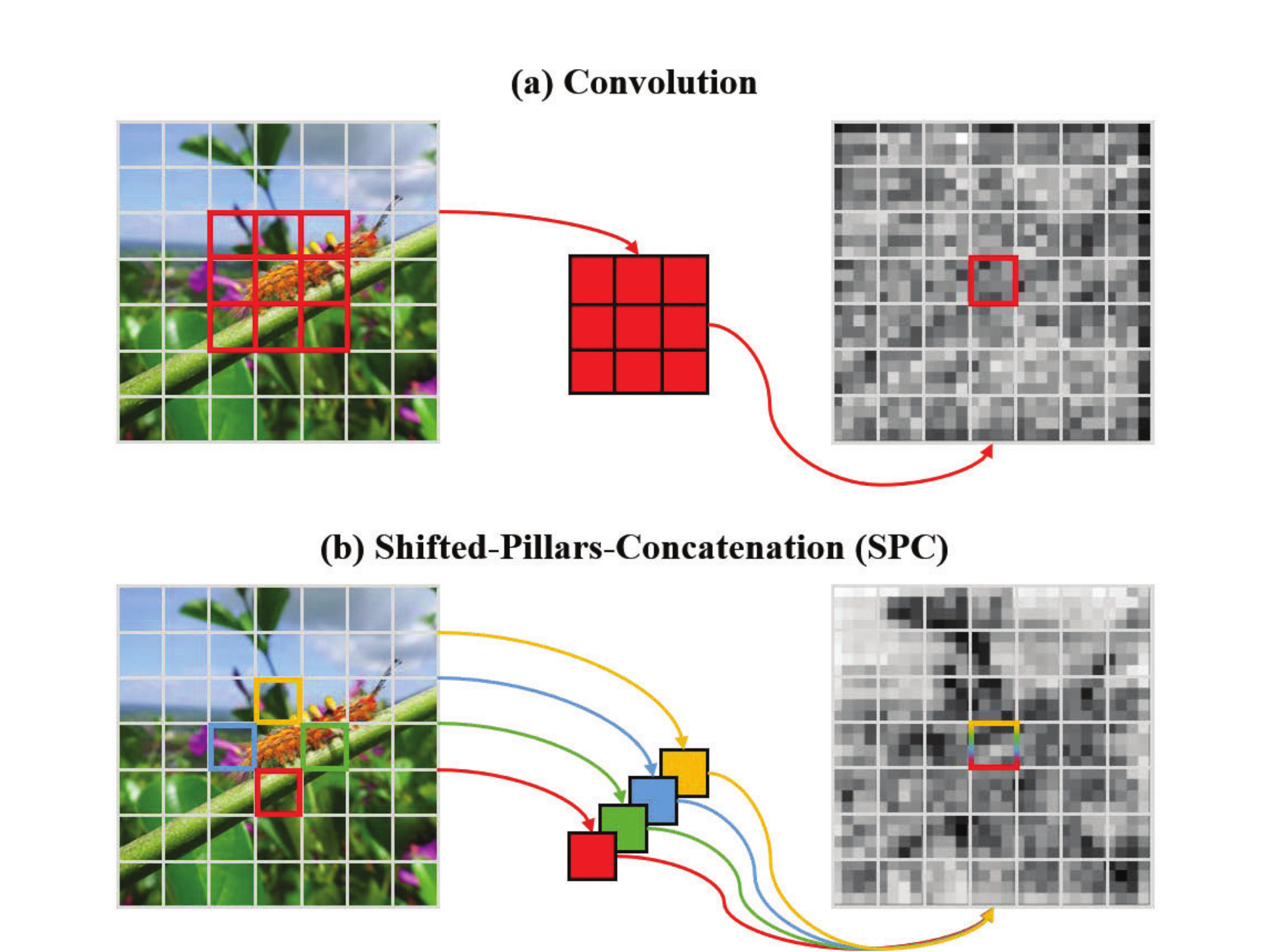}
   \caption{(a) The convolutional layer sequentially slides a local window across each pillar (token) with a larger receptive field (i.e., the colored border), leading to low parallel computation and redundant representation.
   (b) The proposed SPC module adopts a window-free strategy. 
   It applies four linear filters which encode the local features for all pillars in parallel from their neighbors of four directions, exploiting the locality elaborately and simultaneously.}
   \label{fig:onecol}
\end{figure}

Deep architectures in computer vision have evolved from Convolutional Neural Networks (CNNs), through Vision Transformers (ViTs), and now to Multi-Layer Perceptrons (MLPs).
CNNs \cite{krizhevsky2017alexnet,simonyan2014vggnet,he2016resnet} primarily utilize \textit{convolution} to aggregate local features but struggle to capture global dependencies between long-range pillars (tokens) in an image. 
ViTs \cite{dosovitskiy2020vit,touvron2021deit} employ \textit{self-attention mechanism} to consider all pillars from a global perspective.
Unfortunately, the \textit{self-attention mechanism} suffers from high computational complexity.
To overcome this weakness, MLP-based models \cite{tolstikhin2021mlp-mixer} replace the self-attention layers with simple MLPs to perform \textit{token(spatial)-mixing} across the input pillars, thereby significantly reducing the computational costs. 
However, early MLP models \cite{tolstikhin2021mlp-mixer, touvron2022resmlp} encounter the challenges in sufficiently incorporating local dependencies. 
As a solution, researchers have proposed hybrid models \cite{tang2022sMLPNet, li2023convmlp} that combine convolutional layers with MLPs to achieve a balance between capturing local and global information, bringing stable performance improvements.\\

Convolutional layers slide a local window across an image to introduce locality and translation-invariance, which have brought great successes for CNNs \cite{krizhevsky2017alexnet, he2016resnet} and also inspired a number of influential ViTs \cite{wu2021cvt,liu2021swin}.
Nevertheless, convolution has inherent drawbacks. 
First, it may introduce redundancy, especially to the edge features. 
The convolution aggregates pixels in a local window with a larger receptive scope, while the edge features, such as \textit{shape} and \textit{contour}, often consist of only a few pixels that cannot fully fill the scope. Therefore, the edges can get mixed information with the background, leading to redundant representation. 
Additionally, the sliding window needs to encode features sequentially and individually at each position. This sequential nature leads to convolution calculations with limited parallel computing capability.\\

\begin{figure*}[t]
    \centering
    \includegraphics[width=1 \linewidth]{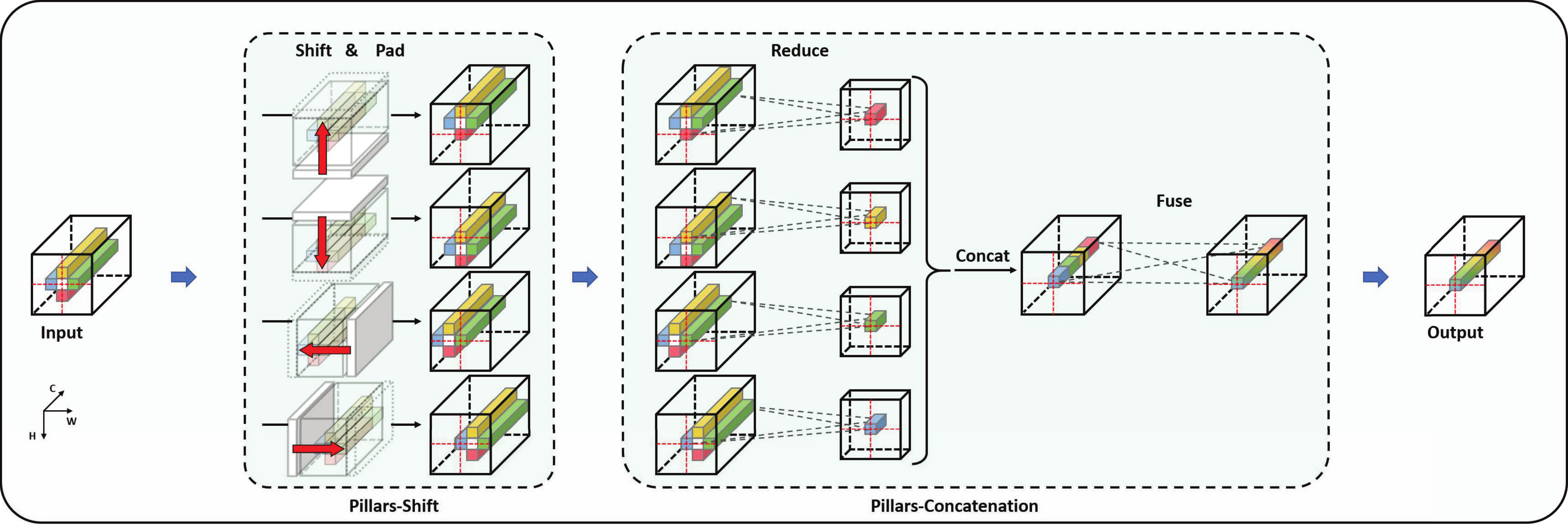}

    \caption{
    The SPC module consists of two processes: Pillars-Shift (\textit{Shift} + \textit{Pad}) and Pillars-Concatenation (\textit{Reduce} + \textit{Concat} + \textit{Fuse}). In Pillars-Shift, the input image is recurrently shifted along four directions to create neighboring maps, while \textit{Pad} is used to maintain the feature size by padding these maps with pillars of specific values. In Pillars-Concatenation, \textit{Reduce} is achieved through four C $\times$ C/4 linear projections, and \textit{Fuse} is accomplished through a C $\times$ C linear projection, where C represents the number of input feature channels.}
    \label{spc_workflow}
\end{figure*}

In this paper, we seek to break the sequential windowing scheme and present an alternative to convolution.
To this end, we propose a new MLP-based module called Shifted-Pillars-Concatenation (SPC), which consists of two processes: Pillars-Shift and Pillars-Concatenation. 
In Pillars-Shift, we shift the input image along four directions (\textit{i.e.,} up, down, left, right) to create four neighboring maps, with the local information for all pillars decomposed into four respective groups according to the orientation of neighboring pillars. 
In Pillars-Concatenation, we apply four linear transformations to individually encode these maps of discrete groups and then concatenate them together, with each pillar achieving the simultaneous and elaborate aggregation of local features from its four neighbors.
% with achieving the simultaneous and elaborate aggregation of local features for each pillar from four neighboring pillars.
Based on the proposed SPC, we introduce a pure-MLP architecture namely Caterpillar, which is built by replacing the (depth-wise) convolutional layer with the proposed SPC module in a Conv-MLP hybrid model (\textit{i.e.}, sMLPNet\cite{tang2022sMLPNet}). The Caterpillar inherits the advantages of sMLPNet, which clearly separates the \textit{local-} and \textit{global-mixing} operations in its spatial-mixing blocks and utilizes the sparse-MLP (sMLP) layer to aggregate global features (Figure \ref{smlp_cprblock}, left), while leveraging the SPC module to exploit locality. \\

For experiments, we uniformly validate the direct application of the Caterpillar with various vision architectures (\textit{i.e.}, CNNs, ViTs, MLPs, Hybrid models) on common-used small-scale images \cite{vinyals2016mini-imagenet,krizhevsky2009cifar,xiao2017fashion-mnist}, among which the Caterpillar achieves the best performance on all used datasets. On the popular ImageNet-1K benchmark, the Caterpillar series attains better or comparable performance to recent state-of-the-art methods (\textit{e.g.}, Caterpillar-B, 83.7\%). Caterpillar also possesses excellent scalability and transfer capability through corresponding experiments.
On the other hand, in all experiments, the Caterpillars obtain higher accuracy than the baseline sMLPNets, while changing the convolution to the SPC module in ResNet-18 brings 4.7\% top-1 accuracy gains on ImageNet-1K dataset, demonstrating the potential of SPC to serve as an alternative to convolution in both plug-and-play and independent ways. \\

In summary, the major contributions of this paper are listed as follows:
\begin{itemize}
  \item  We propose a novel SPC module, which adopts a window-free scheme and can exploit local information more elaborately and simultaneously than convolution.
 
  \item We introduce a new pure-MLP model called Caterpillar, which utilizes SPC and sMLP module to aggregate the local and global information in a sequential way.

  \item Extensive experiments demonstrate the excellent scalability, transfer capability, as well as classification performance of the Caterpillar on both small- and large-scale image recognition tasks, with better performance of SPC than convolution.
  
\end{itemize}

\section{Related Work}
\label{sec:relatedwork}

\subsection{Local Modeling Approaches}

The idea of local modeling can be traced back to research on the organization of the visual cortex \cite{hubel1962receptive,hubel1965receptive}, which inspired Fukushima to introduce the Cognitron \cite{fukushima1975cognitron}, a neural architecture that models nearby features in local regions. 
Departing from biological inspiration, Fukushima further proposed Neocognitron \cite{fukushima1980neocognitron}, which introduces weight sharing across spatial locations through a sliding window strategy. 
% Following the Neocognitron, 
LeCun combined weight sharing with back-propagation algorithm and introduced LeNet \cite{lecun1989conv1,lecun1989conv2,lecun1998lenet}, laying the foundation for the widespread adoption of CNNs in the Deep Learning era. 
Since 2012, when AlexNet \cite{krizhevsky2017alexnet} achieved remarkable performance in the ImageNet classification competition, convolution-based methods have dominated the field of computer vision for nearly a decade.
With the popularity of CNNs, research efforts have been devoted to improving individual convolutional layers, such as depth-wise convolution \cite{xie2017dwconv} and deformable convolution \cite{dai2017deformableConv}. 
On the other hand, alternative approaches to replace convolution have also been explored, such as the \textit{shift}-based methods involving sparse-shift \cite{chen2019sparseshift} and partial-shift \cite{lin2019paricalshift}.
The idea behind these approaches is to move each channel of the input image in different spatial directions, and mix spatial information through linear transformations across channels.
The proposed SPC module also builds upon the \textit{shift} idea but shifts the entire image into four neighboring maps in the process of Pillars-Shift, while making use of the linear projections and concatenation in Pillars-Concatenation.

\subsection{Neural Architectures for Vision}

\noindent \textbf{CNNs and Vision Transformers.} 
CNNs have achieved remarkable success in computer vision, with well-known models including AlexNet \cite{krizhevsky2017alexnet}, VGG \cite{simonyan2014vggnet} and ResNet \cite{he2016resnet}. 
The attention-based Transformer, initially proposed for machine translation \cite{vaswani2017transformer}, has been successfully applied to vision tasks with the introduction of Vision Transformer (ViT) \cite{dosovitskiy2020vit}.
Since then, various advancements have been proposed to improve training efficiency and model performance for ViTs, such as data-efficient training strategy \cite{touvron2021deit} and pyramid architecture \cite{wang2021pvt,liu2021swin}, which have also benefited the entire vision field. 
At the core of Transformer models lies the \textit{multi-head self-attention mechanism}. 
The proposed SPC module shares the similar operation to the \textit{multi-head} settings, as it encodes local neighboring information from different representation subspaces with multiple linear filters in the Pillars-Concatenation process. \\

\noindent\textbf{Vision MLPs.} 
Vision MLPs \cite{tolstikhin2021mlp-mixer, touvron2022resmlp, chen2023cyclemlp, guo2022hiremlp, tang2022wavemlp, hou2022vip, wang2022dynamixer, yu2022s2mlp, yu2021s2mlpv2, lian2021asmlp} have also made significant progress since the invention of MLP-Mixer \cite{tolstikhin2021mlp-mixer}, which alternatively conducts the \textit{token-mixing} (cross-location) operations and \textit{channel-mixing} (per-location) operations to aggregate spatial and channel information, respectively.
Early MLP-based models, such as MLP-Mixer \cite{tolstikhin2021mlp-mixer} and ResMLP \cite{touvron2022resmlp}, perform \textit{token-mixing} across all pillars from a global perspective, lacking the ability to effectively model local features. 
As a result, a number of studies propose to enhance MLPs with local modeling capabilities.
Hire-MLP \cite{guo2022hiremlp}, for instance, performs \textit{Inner-} and \textit{Cross-region Rearrangement} to encode the local and global information in parallel, while AS-MLP \cite{lian2021asmlp} adopts an \textit{axial-shift strategy} that shifts each channel of the image along two directions. 
Our Caterpillar is built with the SPC module, which shifts the entire image into four neighboring maps, enabling the elaborate and simultaneous encoding of local information for all pillars. \\

\noindent\textbf{Hybrid Architectures.} 
Apart from the pure-MLP methods\cite{guo2022hiremlp}, which capture both local and global dependencies fully in MLP-based approaches, there have been developments in combining MLPs with convolutional layers to separately aggregate these two types of information \cite{li2023convmlp, tang2022sMLPNet, cao2023stripmlp}.
% MLPs with convolutional layers to separately aggregate these two types of information have also been developed \cite{li2023convmlp, tang2022sMLPNet, cao2023stripmlp}.
Among them, sMLPNet \cite{tang2022sMLPNet} introduces a \textit{sparse-MLP module} to aggregate global information while using the \textit{depth-wise convolutional (DWConv) layer} to model local features. 
Concurrent with our work, Strip-MLP \cite{cao2023stripmlp} chiefly replaces the \textit{sparse-MLP} (\textit{i.e.,} the global-mixing module) in sMLPNet with a \textit{Strip-MLP layer}, achieving superior scores on both large- and small-scale image datasets.
The proposed Caterpillar is also built upon the sMLPNet but replaces the \textit{DWConv} (\textit{i.e.,} the local-mixing module) with the SPC module, resulting in a pure-MLP architecture, which also attains excellent performance on various-scale image recognition tasks.

\section{Method}
\label{sec:method}

\subsection{Shifted-Pillars-Concatenation Module}

In this section, we first introduce the SPC module, of which the working procedure can be decomposed into Pillars-Shift and Pillars-Concatenation, as shown in Figure \ref{spc_workflow}. 
Then, we analyze its computational parameters with that of the standard and depth-wise convolutional layers. \\

% \subsubsection{Shifted-Pillars-Concatenation}

\noindent \textbf{3.1.1 Shifted-Pillars-Concatenation}

\noindent \textbf{Pillars-Shift.} This process is to shift and pad an input image into four neighboring maps, which can be formulated as:
\begin{eqnarray}
\begin{array}{cc}
\mathrm{PS} \left ( \mathbf{X} \mid  dir, s, p_{m} \right ) =  \mathrm{Pad} \left ( \mathrm{Shift} \left ( \mathbf{X}, dir, s \right )  ,  p_{m}  \right ), & 
dir \subseteq \mathcal{D}_s,  
\end{array}
\end{eqnarray}

where $\mathbf{X}$ is an image, $dir$, $s$ and $p_{m}$ denote the shifting direction, shifting steps and padding mode, respectively. $\mathcal{D}_s$ is a set containing shifting directions. 

Specifically, let $\mathbf{x}_{ij} \in \mathbb{R}^{C} $ denote a feature vector (referred to as "pillar" and also depicted as pillars in Figure \ref{spc_workflow}, so as to clearly and accurately express and visualize the workflow in the SPC module), we can have an image:
% (to differentiate it from)
% (\textcolor{red}{referred to as "pillar", to distinguish from the widely-used word of "token" which has broader usages and meanings, \textit{e.g.,} “token” \textit{vs} “patch”, “token-mixing” \textit{vs} “spatial-mixing”}), we can have an image:
% Specifically, let $\mathbf{x}_{ij} \in \mathbb{R}^{C} $ denote a feature vector called a "pillar" (to differentiate it from)
% (\textcolor{red}{referred to as "pillar", to distinguish from the widely-used word of "token" which has broader usages and meanings, \textit{e.g.,} “token” \textit{vs} “patch”, “token-mixing” \textit{vs} “spatial-mixing”}), we can have an image:
\begin{equation*}
\mathbf{X}_{in} = 
\left [
\begin{matrix}
\mathbf{x}_{11} & \mathbf{x}_{12} & \cdots & \mathbf{x}_{1W} \\\
\mathbf{x}_{21} & \mathbf{x}_{22} & \cdots & \mathbf{x}_{2W} \\\
\vdots & \vdots & \vdots & \vdots \\\
\mathbf{x}_{(H-1)1} & \mathbf{x}_{(H-1)2} & \cdots & \mathbf{x}_{(H-1)W} \\\
\mathbf{x}_{H1} & \mathbf{x}_{H2} & \cdots & \mathbf{x}_{HW} 
\end{matrix} \right ],
\end{equation*}
which means $\mathbf{X}_{in}\in \mathbb{R} ^{H \times W\times C}$, where $H$, $W$ and $C$ represent the width, height and channel number, respectively. 
Taking $\mathbf{X}_{u}$ (\textit{i.e.,} the up-wise neighboring map) as an example, we first perform a \textit{shift} operation on $\mathbf{X}_{in}$ by setting $dir$=$`up'$ and $s=1$, so that $\mathbf{X}_{in}$ is transformed to:
\begin{equation*}
\mathbf{X}^{'}_{\mathit{u}} = 
\left [
\begin{matrix}
\mathbf{x}_{21} & \mathbf{x}_{22} & \cdots & \mathbf{x}_{2W} \\\
\vdots & \vdots & \vdots & \vdots \\\
\mathbf{x}_{(H-1)1} & \mathbf{x}_{(H-1)2} & \cdots & \mathbf{x}_{(H-1)W} \\\
\mathbf{x}_{H1} & \mathbf{x}_{H2} & \cdots & \mathbf{x}_{HW} \\\
\end{matrix} \right ],
\end{equation*}
where $\mathbf{X}'_{\mathit{u}}\in \mathbb{R}^{(H-1) \times W\times C}$. Then, we \textit{pad} $\mathbf{X}^{'}_{\mathit{u}}$ according to the Zero Padding and attain the $\mathbf{X}_{u}$:
\begin{equation*}
\mathbf{X}_{\mathit{u}} = 
\left [
\begin{matrix}
\mathbf{x}_{21} & \mathbf{x}_{22} & \cdots & \mathbf{x}_{2W} \\\
\vdots & \vdots & \vdots \\\
\mathbf{x}_{(H-1)1} & \mathbf{x}_{(H-1)2} & \cdots & \mathbf{x}_{(H-1)W} \\\
\mathbf{x}_{H1} & \mathbf{x}_{H2} & \cdots & \mathbf{x}_{HW} \\\
\mathbf{0} & \mathbf{0} & \cdots & \mathbf{0}
\end{matrix} \right ],
\end{equation*}
where $\mathbf{X}_{\mathit{u}}\in \mathbb{R}^{H \times W\times C} $. 
By default settings with $\mathcal{D}_s$ of $ [ `up', `down', $ $  `left', `right' ] $, the input $\mathbf{X}_{in}$ will be transformed into four neighboring maps of $\mathbf{X}_{u}, \mathbf{X}_{d}, \mathbf{X}_{l}, \mathbf{X}_{r}$, where $\mathbf{X}_{\mathit{d}}, \mathbf{X}_{\mathit{l}}, \mathbf{X}_{\mathit{r}} \in \mathbb{R}^{H \times W\times C} $.\\

\noindent\textbf{Pillars-Concatenation.} Obviously, Pillars-Shift has no parameter learning, which would weaken the representation capability of the module. 
To overcome this deficiency, we introduce the Pillars-Concatenation process. 
Specifically, the neighboring maps $\mathbf{X}_{u}, \mathbf{X}_{d}, \mathbf{X}_{l}, \mathbf{X}_{r}$ are projected through four independent fully-
connected (FC) layers. 
The parameters are $\mathbf{W}_{u}, \mathbf{W}_{d}, \mathbf{W}_{l}, \mathbf{W}_{r} \in \mathbb{R}^{C \times C/4}$, respectively, so as to reduce the number of neighboring maps' channels into $C/4$. 
After that, all of the reduced maps are concatenated along the channel dimension and then projected again, by an FC layer with the parameters $\mathbf{W}\in \mathbb{R} ^{C \times C}$ to fuse the local features. 
This process can be represented as:
\begin{eqnarray}
         \mathrm{PC} \left ( \mathbf{X} \right ) =  \mathrm{Concat} \left ( \mathbf{X}_{u}\mathbf{W}_u,\mathbf{X}_{d}\mathbf{W}_d,\mathbf{X}_{l}\mathbf{W}_l, \mathbf{X}_{r}\mathbf{W}_r \right )\mathbf{W},
\end{eqnarray}

Through the Pillars-Concatenation, the four neighboring maps are reduced, concatenated and finally fused into the $\mathbf{X}_{out}\in \mathbb{R} ^{H \times W\times C}$, with the local information for all pillars within the image aggregated in parallel.\\

% \subsubsection{Parameter Analysis with Convolutions}
\noindent \textbf{3.1.2 Parameter Analysis with Convolutions}

For an input image with the input dimension of $d_{in}$ and output dimension of $d_{out}$, the number of parameters in standard
 and depth-wise convolution (DWConv) can be calculated as $d_{in} \times k^{2} \times d_{out}$ and  $d_{in} \times k^{2}$, respectively, with $k$ representing the kernel size. 
In comparison, The parameters of SPC module are $d_{in} \times d_{in} + d_{in} \times d_{out}$ (detailed as $d_{in} \times d_{in}/4 \times 4 + d_{in} \times d_{out}$).
In the typical scenario, where $d_{in}$ is equal to $d_{out}$, the parameters of a standard 3 $\times$ 3 convolution are $9 \times {d_{in}}^{2}$, which is 4.5 times larger than that of SPC, \textit{i.e.}, $ 2\times {d_{in}}^{2}$, demonstrating that the SPC module has lower computational complexity than the standard convolutional layers. Additionally, the depth-wise settings reduce the parameters in DWConv into $9 \times {d_{in}}$, which is lower than SPC and might inspired future works that improve SPC through such lightweight techniques.

% \textcolor{red}{Combining convolution with the depth-wise setting (\textit{i.e.}, 3 $\times$ 3 depth-wise convolution adapted in sMLPNet) can reduce the parameters into $9 \times d_{in}$, which offers a promising avenue for enhancing the lightweight nature of SPC by leveraging techniques such as depth-wise (et.al.).}
% providing an future improvement for SPC 
% to become more lightweight through the techniques like depth-wise setting.

% \textcolor{red}{Additionally, the parameters in the 3 $\times$ 3 depth-wise convolution (adapted in sMLPNet) are $9 \times d_{in}$.} 

\subsection{Caterpillar Block}

\begin{figure}[t]
  \centering
   \includegraphics[width=0.8 \linewidth]{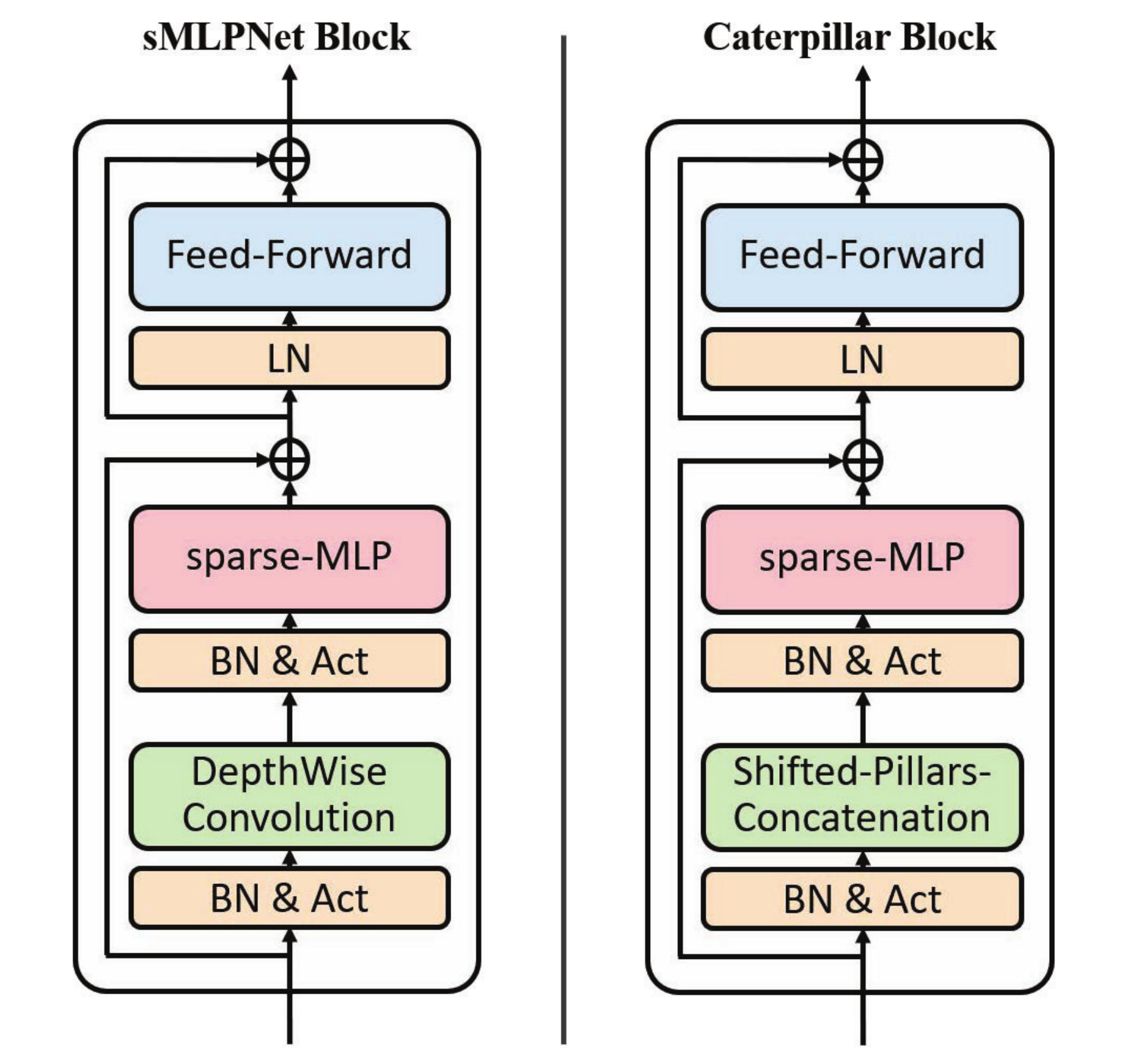}

   \caption{The structures of sMLPNet and Caterpillar blocks.}
   \label{smlp_cprblock}
\end{figure}

Caterpillar block is built by replacing the depth-wise convolution with the SPC module in sMLPNet block \cite{tang2022sMLPNet}, in which a sparse-MLP (sMLP) module (illustrated in Appendix \ref{sup:sMLP_module}) is introduced for aggregating global features. 
As illustrated in Figure \ref{smlp_cprblock}, a Caterpillar block contains three basic modules: an SPC module and an sMLP module, with a BatchNorm (BN) and a GELU nonlinearity applied before them, and a FFN module, which follows a LayerNorm (LN) layer.
% \textcolor{red}{with following a LayerNorm (LN) layer. }
The SPC and sMLP form the token-mixing component and the FFN servers as channel-mixing module, with applied two residual connections.

Given an image $\mathbf{X}\in \mathbb{R} ^{H \times W\times C}$, the calculation in the Caterpillar block can be formulated as:
\begin{eqnarray}
         \mathbf{X}'  =  \mathrm{SPC} \left (\mathrm{GELU} \left ( \mathrm{BN} \left (\mathbf{X} \right ) \right ) \right ),
\end{eqnarray}
\begin{eqnarray}
         \mathbf{Y}  = \mathrm{sMLP} \left ( \mathrm{GELU} \left ( \mathrm{BN} \left ( \mathbf{X}' \right ) \right ) \right ) + \mathbf{X},
\end{eqnarray}
\begin{eqnarray}
         \mathbf{Z}  = \mathrm{FFN} \left (\mathrm{LN} \left ( \mathbf{Y} \right ) \right ) + \mathbf{Y},
\end{eqnarray}
where $\mathbf{X}'$ denotes the output features of the SPC layer, $\mathbf{Y}$ and $\mathbf{Z}$ represent the output of token-mixing and channel-mixing modules, respectively. 

\subsection{Caterpillar Architectures}
\label{Section 3.3}

We build the Caterpillar architectures in a pyramid structure of four stages, which first represent the input images into patch-level features, and gradually shrink the spatial size of the feature maps as the network deepens. 
This enables Caterpillar to leverage the scale-invariant property of images as well as make full use of multi-scale features.

We introduce five variants of the Caterpillar architectures, \textit{i.e.}, Caterpillar-Mi, -Tx, -T, -S and -B, with different numbers of Caterpillar blocks stacked in their four stages. The architecture hyper-parameters of these variants are:

% \begin{itemize}[leftmargin=0.6cm]
\begin{itemize}
    \item Caterpillar-Mi: $C$ = 40,  layer numbers = {\{2,6,10,2\}}

    \item Caterpillar-Tx: $C$ = 60,  layer numbers = {\{2,8,14,2\}}

    \item Caterpillar-T:  $C$ = 80,  layer numbers = {\{2,8,14,2\}}

    \item Caterpillar-S:  $C$ = 96,  layer numbers = {\{2,10,24,2\}}

    \item Caterpillar-B:  $C$ = 112,  layer numbers = {\{2,10,24,2\}}
    
\end{itemize}
where $C$ is the channel number of the hidden layers in the first stage, and layer numbers denote the number of blocks in each of their four stages. Detailed configurations are in Appendix \ref{sup:detailArch}.

\section{Experiments}
\label{sec:experiments}

% \textcolor{red}{In this Section, we organize the experiments as follows:
% The experiments are organized as follows:
We compare the classification performance of Caterpillar with various vision models on small-scale images as well as ImageNet-1k dataset in Sec. \ref{Section 4.1}, \ref{Section 4.5} and Sec. \ref{Section 4.2}, respectively, with its scalability and transfer capability also verified in Sec. \ref{Section 4.1} and Sec. \ref{Section 4.5}.
Then, we conduct ablation studies in Sec. \ref{Section 4.3} to find the best settings for the proposed SPC module and Caterpillar architecture.
Additionally, the comparison between sMLPNet and Caterpillar in Sec. \ref{Section 4.1}, \ref{Section 4.2} as well as the experiments in Sec. \ref{Section 4.6} is to verify SPC to be an alternative to convolution in both plug-and-play and independent manners, with visualization analysis for SPC provided in Sec. \ref{Section 4.4}.

\subsection{Small-scale Image Classification}
\label{Section 4.1}

\noindent \textbf{Datasets.} We conduct small-scale image recognition experiments on four commonly-used benchmarks: Mini-ImageNet (MIN) \cite{vinyals2016mini-imagenet}, CIFAR-10 (C10) \cite{krizhevsky2009cifar}, CIFAR-100 (C100) \cite{krizhevsky2009cifar} and Fashion-MNIST (Fashion) \cite{xiao2017fashion-mnist}. 
We utilize these images in their original sizes, different from the settings in \cite{cao2023stripmlp} that resized images into 224 $\times$ 224. 

\noindent \textbf{Experimental Settings.} We evaluate the Caterpillar with fourteen representative vision models, including six MLP models \cite{touvron2022resmlp,chen2023cyclemlp,guo2022hiremlp,tang2022wavemlp,hou2022vip,wang2022dynamixer}, two CNN models \cite{he2016resnet, trockman2022convmixer}, three Transformer models \cite{touvron2021deit, liu2021swin,zhang2022nest}, and three hybrid models \cite{hassani2021cct,tang2022sMLPNet,cao2023stripmlp}, as tagged in Table \ref{tab-result-M1}.
All models are directly trained on the small images without extra data.
To enable the models adaptable to small-sized images (\textit{e.g.}, 32 $\times$ 32), we change their \textit{patch embedding layers} into small \textit{patch sizes} according to uniform rules, of which the detailed implementation is provided in Appendix \ref{sup:Imp_SmallImage}.
For a fair comparison, we adopt the same training strategies that were presented in their original papers (for ImageNet-1K), which are presented in Appendix \ref{sup:trainingStrategies}.

\begin{table}[h]
\caption{Results  (\%) of Caterpillar and other MLP / CNN / Transformer / Hybrid vision models on four small-scale datasets. 
As the model parameters and FLOPs are similar on these datasets, we just report those metrics on CIFAR-10 for clarity. 
% The result of CCT-7/3×1 on MIN is omitted, since the FLOPs of it (\textit{i.e.}, 6.6G) is much larger than that of other models. 
The Caterpillar-T$^{\dag}$ scales the number of channels to [72, 144, 288, 576], with similar computational costs to the sMLPNet-T. \textcolor{blue}{$\blacktriangle$} CNN, \textcolor{green}{$\blacklozenge$} Transformer, 
\textcolor{orange}{$\blacksquare$} MLP, \textcolor{pink}{$\blacksquare$} Hybrid, \textcolor{red}{$\bigstar$} Ours.}
  \label{tab-result-M1}
  \centering
  \renewcommand\tabcolsep{3.0pt}
    \scalebox{0.85}{
  % \begin{tabular}{l|c|c|c|c|c|c|c}
  \begin{tabular}{l|cccc|cc}
    \arrayrulecolor{black}\toprule
    Networks                 & MIN                  & C10                 & C100                & Fashion            & Params   & FLOPs \\
    \midrule
    % \midrule
    \textcolor{green}{$\blacklozenge$} DeiT-Ti\cite{touvron2021deit}                   & 54.55                & 88.87               & 67.46               & 92.97                & 5.4M              & 0.3G \\
    \textcolor{green}{$\blacklozenge$} NesT-T\cite{zhang2022nest}                   & \underline{73.44}    & \underline{94.05}   & \underline{75.60}   & \underline{94.26}    & 6.4M              & 2.3G \\
    % \arrayrulecolor{gray}\midrule
    \textcolor{pink}{$\blacksquare$} CCT-7/3×1\cite{hassani2021cct}                &   --                  & 91.80   & 74.09   & 93.70    & 3.7M              & 0.9G \\
    % \arrayrulecolor{gray}\midrule
    \textcolor{red}{$\bigstar$} \textbf{Caterpillar-Mi}          & \textbf{74.14}       & \textbf{95.54}      & \textbf{79.41}      & \textbf{95.14}       & 5.9M              & 0.4G \\
    % \midrule
    \midrule
    \textcolor{blue}{$\blacktriangle$} ResNet-18\cite{touvron2022resmlp}                & 70.95    & 95.54   & 77.66   & 95.11    & 11.2M             & 0.7G \\
    \textcolor{blue}{$\blacktriangle$} ConvMixer\_768/32\cite{trockman2022convmixer}       & 57.94                & 91.54    & 70.13    & 93.36     & 19.4M   & 1.2G \\      
    \textcolor{orange}{$\blacksquare$} ResMLP-S12\cite{touvron2022resmlp}               & 68.63                & 93.67   & 76.44   & 94.58    & 14.3M             & 0.9G \\
    \textcolor{orange}{$\blacksquare$} CycleMLP-B1\cite{chen2023cyclemlp}             & 70.68                & 88.06    & 66.17    & 92.87     & 12.7M   & 0.1G \\
    \textcolor{orange}{$\blacksquare$} HireMLP-Tiny\cite{guo2022hiremlp}              & 71.66                & 86.42    & 62.13    & 92.35     & 17.6M   & 0.1G \\
    \textcolor{orange}{$\blacksquare$} Wave-MLP-T\cite{tang2022wavemlp}               & 72.15    & 88.85    & 65.92    & 92.83     & 16.7M   & 0.1G \\
    % \arrayrulecolor{gray}\midrule
    % \arrayrulecolor{gray}\midrule
    \textcolor{pink}{$\blacksquare$} Strip-MLP-T*\cite{cao2023stripmlp}               & \underline{76.05}    & \underline{96.34}   & \underline{82.53}   & \underline{95.33}    & 16.3M             & 0.8G \\
    \textcolor{red}{$\bigstar$} \textbf{Caterpillar-Tx}          & \textbf{77.27}       & \textbf{96.54}      & \textbf{82.69}      & \textbf{95.38}       & 16.0M             & 1.1G \\
    % \midrule
    \midrule
    \textcolor{blue}{$\blacktriangle$} ResNet-34\cite{he2016resnet}                & 72.03    & 95.92   & 79.53   & 95.48                & 21.3M             & 1.5G \\
    \textcolor{blue}{$\blacktriangle$} ResNet-50\cite{he2016resnet}                & 72.65    & 96.06   & 79.11   & 95.28                & 23.7M             & 1.6G \\
    % \arrayrulecolor{gray}\midrule
    \textcolor{green}{$\blacklozenge$} DeiT-S\cite{touvron2021deit}                   & 42.41                & 83.10               & 64.65               & 93.43    & 21.4M             & 1.4G \\
    \textcolor{green}{$\blacklozenge$} Swin-T\cite{liu2021swin}                   & 53.11    & 85.69   & 67.60   & 89.90                & 27.6M             & 1.4G \\
    % \arrayrulecolor{gray}\midrule
    \textcolor{orange}{$\blacksquare$} ResMLP-S24\cite{touvron2022resmlp}               & 69.63                & 94.76               & 78.65   & 95.27    & 28.5M             & 1.9G \\
    \textcolor{orange}{$\blacksquare$} CycleMLP-B2\cite{chen2023cyclemlp}             & 71.11                & 88.84               & 67.83               & 93.41                & 22.6M             & 0.1G \\
    \textcolor{orange}{$\blacksquare$} HireMLP-Small\cite{guo2022hiremlp}               & 73.86    & 88.51               & 62.54               & 92.70                & 32.6M             & 0.1G \\
    \textcolor{orange}{$\blacksquare$} Wave-MLP-S\cite{tang2022wavemlp}               & 67.51                & 88.37               & 63.24               & 92.96                & 30.2M             & 0.1G \\
    \textcolor{orange}{$\blacksquare$} ViP-Small/7\cite{hou2022vip}                  & 70.94                & 94.12               & 78.28               & 95.22                & 24.7M             & 1.7G \\
    \textcolor{orange}{$\blacksquare$} DynaMixer-S\cite{wang2022dynamixer}             & 71.40                & 95.32   & 78.34               & 95.14                & 25.2M             & 1.8G \\
    % \arrayrulecolor{gray}\midrule
    \rowcolor{blue!8}\textcolor{pink}{$\blacksquare$} sMLPNet-T\cite{tang2022sMLPNet}               & 77.07    & 96.87   & 82.89   & 95.53    & 23.5M             & 1.6G \\
    \textcolor{pink}{$\blacksquare$} Strip-MLP-T\cite{cao2023stripmlp}               & 76.47    & 96.48   & 82.59   & 95.50    & 22.5M             & 1.2G \\
    % \arrayrulecolor{gray}\midrule
    \rowcolor{blue!8}\textcolor{red}{$\bigstar$} \textbf{Caterpillar-T$^{\dag}$}           & \underline{77.56}       & \underline{97.08}      & \underline{83.12}      & \underline{95.57}       & 23.0M             & 1.6G \\
    \rowcolor{blue!8}\textcolor{red}{$\bigstar$} \textbf{Caterpillar-T}           & \textbf{78.16}       & \textbf{97.10}      & \textbf{83.86}      & \textbf{95.72}       & 28.4M             & 1.9G \\
    % \midrule
    \midrule
    \textcolor{red}{$\bigstar$} \textbf{Caterpillar-S}           & 78.94       & 97.22               & 84.40               & 95.80                & 58.0M             & 4.1G \\
    \textcolor{red}{$\bigstar$} \textbf{Caterpillar-B}           & \textbf{79.06}                & \textbf{97.35}      & \textbf{84.77}      & \textbf{95.85}       & 78.8M             & 5.5G \\
    \arrayrulecolor{black}\bottomrule
  \end{tabular}
  }
\end{table}

\noindent \textbf{Results.}  Table \ref{tab-result-M1} presents the classification results of different methods on the four small-scale image datasets. As we can see, the proposed Caterpillar outperforms the sMLPNet on all the four benchmarks (\textit{e.g.,} Caterpillar-T$^{\dag}$, 23M, 77.56\% \textit{vs} sMLPNet-T, 24M, 77.07\% on MIN), showcasing the better classification capability of the SPC than convolutional layers and its potential to be an alternative to convolution in plug-and-play ways.
Additionally, the Caterpillar attain the best scores among all tested architectures, \textit{i.e.,} the Caterpillar-T reaches 78.16\% accuracy on MIN, 97.10\% on C10, 84.86\% on C100, and 95.72\% on Fashion, making it an effective tool for small-scale image recognition tasks.

\noindent \textbf{Scalability analysis.} “Simple algorithms that scale well are the core of deep learning” \cite{he2022mae}. 
Thus, we scale the Caterpillar from -Mi with FLOPs about 0.4G to -B about 5.5G, \textit{i.e.}, Caterpillar-Mi, -Tx, -T, -S and -B. 
It is credible that Caterpillar exhibits excellent scalability on small-scale datasets, as it obtains steady improvement from bigger models.

\begin{table}[t]
  \caption{Results (\%) of Caterpillar and other vision models on ImageNet-1K datasets. \textcolor{blue}{$\blacktriangle$} CNN, \textcolor{green}{$\blacklozenge$} Transformer, \textcolor{orange}{$\blacksquare$} MLP, \textcolor{pink}{$\blacksquare$} Hybrid, \textcolor{red}{$\bigstar$} Ours.}
  \label{tab-M3}
  \centering
  \renewcommand\tabcolsep{10.0pt}
  \scalebox{0.93}{
  \begin{tabular}{l|c|c|c}
    \toprule
    Networks                                         & Params    & FLOPs      & Top-1     \\
    \midrule
    \textcolor{green}{$\blacklozenge$} DeiT-Ti\cite{touvron2021deit}                    & 5M     & 1.1G    & 72.2    \\
    
    \textcolor{orange}{$\blacksquare$} gMLP-Ti\cite{liu2021gmlp}                        & 6M     & 1.4G    & \underline{72.3}    \\
    \textcolor{red}{$\bigstar$} \textbf{Caterpillar-Mi}                                  & 6M     & 1.2G    & \textbf{76.3}    \\
    \midrule
    \textcolor{blue}{$\blacktriangle$} ResNet-18\cite{he2016resnet,wightman2021resnetstrike}         & 12M    & 1.8G    & 70.6    \\
    % \textcolor{green}{$\blacklozenge$} PVT-Tiny\cite{wang2021pvt}                       & 13M    & 1.9G    & 75.1    \\
    \textcolor{orange}{$\blacksquare$} ResMLP-S12\cite{touvron2022resmlp}               & 15M    & 3.0G    & 76.6    \\
    \textcolor{orange}{$\blacksquare$} Hire-MLP-Ti\cite{guo2022hiremlp}               & 18M    & 2.1G    & 79.7    \\
    \textcolor{orange}{$\blacksquare$} Wave-MLP-T\cite{tang2022wavemlp}               & 17M    & 2.4G    & 80.6    \\
    \textcolor{pink}{$\blacksquare$} Strip-MLP-T*\cite{cao2023stripmlp}                  & 18M    & 2.5G    & \textbf{81.2}    \\
    \textcolor{red}{$\bigstar$} \textbf{Caterpillar-Tx}                                  & 16M    & 3.4G    & \underline{80.9}    \\
    \midrule
    \textcolor{blue}{$\blacktriangle$} ResNet-50\cite{he2016resnet,wightman2021resnetstrike}         & 26M    & 4.1G    & 79.8    \\
    \textcolor{blue}{$\blacktriangle$} RegNetY-4G\cite{radosavovic2020regnet}         & 21M    & 4.0G    & 80.0    \\
    \textcolor{green}{$\blacklozenge$} DeiT-S\cite{touvron2021deit}                     & 22M    & 4.6G    & 79.8    \\
    % \textcolor{green}{$\blacklozenge$} PVT-Small\cite{wang2021pvt}                      & 25M    & 3.8G    & 81.2    \\
    \textcolor{green}{$\blacklozenge$} Swin-T\cite{liu2021swin}                         & 29M    & 4.5G    & 81.3   \\
    \textcolor{orange}{$\blacksquare$} ResMLP-S24\cite{touvron2022resmlp}               & 30M    & 6.0G    & 79.4    \\
    \textcolor{orange}{$\blacksquare$} ViP-Small/7\cite{hou2022vip}               & 25M    & 6.9G    & 81.5    \\
    \textcolor{orange}{$\blacksquare$} AS-MLP-T\cite{lian2021asmlp}               & 28M    & 4.4G    & 81.3    \\
    \textcolor{orange}{$\blacksquare$} Hire-MLP-S\cite{guo2022hiremlp}               & 33M    & 4.2G    & 82.1    \\
    \textcolor{orange}{$\blacksquare$} Wave-MLP-S\cite{tang2022wavemlp}               & 30M    & 4.5G    & \textbf{82.6}    \\
    % \textcolor{pink}{$\blacksquare$} CCT-14/7$\times$2\cite{hassani2021cct}             & 22M    & 5.5G    & 80.7    \\
    \rowcolor{blue!8}\textcolor{pink}{$\blacksquare$} sMLPNet-T\cite{tang2022sMLPNet}                  & 24M    & 5.0G    & 81.9    \\
    \textcolor{pink}{$\blacksquare$} Strip-MLP-T\cite{cao2023stripmlp}                  & 25M    & 3.7G    & 82.2    \\
    \rowcolor{blue!8}\textcolor{red}{$\bigstar$} \textbf{Caterpillar-T}                                   & 29M    & 6.0G    & \underline{82.4}    \\
    \midrule
    \textcolor{blue}{$\blacktriangle$} ResNet-101\cite{he2016resnet,wightman2021resnetstrike}         & 45M    & 7.9G    & 81.3    \\
    \textcolor{blue}{$\blacktriangle$} RegNetY-8G\cite{radosavovic2020regnet}         & 39M    & 8.0G    & 81.7    \\
    \textcolor{green}{$\blacklozenge$} Swin-S\cite{liu2021swin}                         & 50M    & 8.7G    & 83.0   \\
    \textcolor{orange}{$\blacksquare$} ViP-Medium/7\cite{hou2022vip}               & 55M    & 16.3G    & 82.7    \\
    \textcolor{orange}{$\blacksquare$} AS-MLP-S\cite{lian2021asmlp}                & 50M    & 8.5G    & 83.1    \\
    \textcolor{orange}{$\blacksquare$} Hire-MLP-B\cite{guo2022hiremlp}               & 58M    & 8.1G    & 83.2    \\
    \textcolor{orange}{$\blacksquare$} Wave-MLP-M\cite{tang2022wavemlp}               & 44M    & 7.9G    & \underline{83.4}    \\
    \rowcolor{blue!8}\textcolor{pink}{$\blacksquare$} sMLPNet-S\cite{tang2022sMLPNet}                  & 49M    & 10.3G    & 83.1    \\
    \textcolor{pink}{$\blacksquare$} Strip-MLP-S\cite{cao2023stripmlp}                  & 44M    & 6.8G    & 83.3    \\
    \rowcolor{blue!8}\textcolor{red}{$\bigstar$} \textbf{Caterpillar-S}                                   & 60M    & 12.5G    & \textbf{83.5}    \\
    \midrule
    \textcolor{blue}{$\blacktriangle$} ResNet-152\cite{he2016resnet,wightman2021resnetstrike}         & 60M    & 11.6G    & 81.8    \\
    \textcolor{blue}{$\blacktriangle$} RegNetY-16G\cite{radosavovic2020regnet}         & 84M    & 16.0G    & 82.9    \\
    \textcolor{green}{$\blacklozenge$} DeiT-B\cite{touvron2021deit}                     & 86M    & 17.5G    & 81.8    \\
    \textcolor{green}{$\blacklozenge$} Swin-B\cite{liu2021swin}                         & 88M    & 15.4G    & 83.5   \\
    \textcolor{orange}{$\blacksquare$} ResMLP-B24\cite{touvron2022resmlp}               & 116M    & 23.0G    & 81.0    \\
    \textcolor{orange}{$\blacksquare$} ViP-Large/7\cite{hou2022vip}                  & 88M    & 24.4G    & 83.2    \\
    \textcolor{orange}{$\blacksquare$} AS-MLP-B\cite{guo2022hiremlp}                 & 88M    & 15.2G    & 83.3    \\
    \textcolor{orange}{$\blacksquare$} Hire-MLP-B\cite{guo2022hiremlp}               & 96M    & 13.4G    & \textbf{83.8}    \\
    \textcolor{orange}{$\blacksquare$} Wave-MLP-B\cite{tang2022wavemlp}               & 63M    & 10.2G    & 83.6    \\
    \rowcolor{blue!8}\textcolor{pink}{$\blacksquare$} sMLPNet-B\cite{tang2022sMLPNet}                  & 66M    & 14.0G    & 83.4    \\
    \textcolor{pink}{$\blacksquare$} Strip-MLP-B\cite{cao2023stripmlp}                  & 57M    & 9.2G    & 83.6    \\
    \rowcolor{blue!8}\textcolor{red}{$\bigstar$} \textbf{Caterpillar-B}                                   & 80M    & 17.0G    & \underline{83.7}    \\
    \bottomrule
  \end{tabular}
  }
\end{table}

\subsection{ImageNet Classification}
\label{Section 4.2}

\noindent \textbf{Datasets.} We test the Caterpillar on the ImageNet-1K benchmark \cite{deng2009imagenet1k}, which consists of 1.28M training and 50K validation images belonging to 1,000 categories. 

\noindent \textbf{Experimental Settings.} We train our models on 8 NVIDIA GeForce RTX 3090 GPUs with gradient accumulation techniques. For training strategies, we employ the AdamW \cite{loshchilov2017adamw} optimizer to train our models for 300 epochs, with a weight decay of 0.05 and a batch size of 1024. 
The learning rate is initially 1e-3 and gradually drops to 1e-5 according to the consine schedule. 
The data augmentation includes Random Augment \cite{cubuk2020randaugment}, Mixup \cite{zhang2017mixup}, Cutmix \cite{yun2019cutmix}, Random Erasing \cite{zhong2020randerase}. 
More details are shown in Appendix \ref{sup:trainingStrategies}. 

\noindent\textbf{Results.} Table \ref{tab-M3} presents the performance of Caterpillar with other well-established methods on the ImageNet-1k benchmark. Similar to the results in Section \ref{Section 4.1}, the Caterpillar models consistently outperform their sMLPNet counterparts, which further emphasizes the superiority of the SPC module over convolutional layers, highlighting its potential as a plug-and-play replacement to convolution.
Furthermore, the Caterpillar series exhibit competitive or even superior performance to state-of-the-art networks. For instance, Caterpillar-B achieves the top-1 accuracy of 83.7\%, which slightly surpasses several representative MLP architectures (\textit{e.g.,} Wave-MLP-B, 83.6\%, AS-MLP-B, 83.3\%, ViP-Large/7, 83.2\%),
verifying the efficacy of Caterpillar in tackling large-scale vision recognition tasks.
% that Caterpillar is also capable of handling large-scale vision recognition tasks.

\subsection{Ablation Study}
\label{Section 4.3}

In this section, we ablate essential design components in the proposed Caterpillar architecture. We use the same datasets and experimental settings as in Section \ref{Section 4.1}. The base architecture is Caterpillar-T.\\

% \subsubsection{Pillars-Shift of SPC}
\noindent \textbf{4.3.1 Pillars-Shift of SPC}

\noindent \textbf{Number of shift directions.} 
This hyper-parameter ($N_{D}$) controls the shifting directions of input images so as to determine the receptive field of SPC on local features.
We experiment with $N_{D}$ values ranging from 4 to 9. 
Among them, 4 represents the scope of four neighboring directions (\textit{up, down, left, and right}), and 5 includes the \textit{center} pillar itself.
8 covers a wider scope, including \textit{up, down, left, right, up-left, up-right, down-left, and down-right} directions. When $N_{D}$ is set to 9, it adds the \textit{center} pillar itself, which is similar to the scope of 3x3 convolution.
From Table \ref{tab-A1NumD}, increasing the number of neighbors can bring redundancy, because more background information can be injected into the target pillar -- the similar drawback existed in convolution.
% the best trade-off between computational costs and accuracy is achieved when $N_{D}=4$. 
This result underscores that the local features can be sufficiently obtained from a 4-scoped receptive field.

\begin{table}[h]
  \caption{Results (\%) on different numbers of shift directions in the Pillars-Shift process. The $C$ for the “$N$ = 9” is adjusted to [81, 162, 324, 648] to ensure the divisibility of $C$ \textit{by} $N$ with the \textit{Reduce Weights} of $\mathbf{W} \in \mathbb{R}^{C \times C/N}$}
  % The model with $N_{D}$ of 9$^{\ast}$ adopts the channel numbers of [81, 162, 324, 648] in its four stages.}
  \label{tab-A1NumD}
  \centering
   \renewcommand\tabcolsep{3.0pt}
  \scalebox{0.9}{
  \begin{tabular}{c|cccc|cc}
    \toprule
    Num. of dir. ($N_{D}$)        & MIN    & C10      & C100    & Fashion    & Params    & FLOPs \\
    \midrule
    4                & \underline{78.16}      & \textbf{97.10}      & \textbf{83.86}      & \textbf{95.72}    & 28.4M     & 1.9G \\
    5                & 78.04    & \underline{96.93}    & 83.55    & \underline{95.63}    & 28.4M     & 1.9G \\
    8                & \textbf{78.19}    & 96.92    & 83.58    & 95.57   & 28.4M     & 1.9G \\
    \ \ 9$^{\ast}$                & 77.92    & 96.82    & \underline{83.60}    & 95.52   & 29.1M     & 2.0G \\
    \bottomrule
  \end{tabular}
  }
\end{table}

\noindent \textbf{Number of shift steps.} The hyper-parameter $s$ determines the range of local features for the Pillars-Shift operation.
When $s$ is set to 0, 1, or 2, the input image is shifted 0, 1, or 2 steps along the corresponding directions, which allows the local information for each pillar to be aggregated from itself (no shifting, lacking local modeling capability), neighboring pillars (with a distance of 1), or distant pillars (with a distance of 2), respectively.
Table \ref{tab-A1Step} displays the results of the proposed method with different numbers of shifting steps. The findings indicate that the best performance can be achieved when $s=1$.

\begin{table}[h]
  \caption{The model accuracy  (\%) on three different shift steps in the Pillars-Shift process.}
  \label{tab-A1Step}
  \centering
  \scalebox{1.0}{
  \begin{tabular}{c|cccc}
    \toprule
    shift steps ($s$)       & MIN      & C10      & C100     & Fashion \\
    \midrule
    0                 & \underline{76.71}    & 95.84    & 81.12    & 95.22   \\
    1                 & \textbf{78.16}      & \textbf{97.10}      & \textbf{83.86}      & \textbf{95.72} \\
    2                 & 76.29    & \underline{96.17}    & \underline{82.04}    & \underline{95.26}   \\
    \bottomrule
  \end{tabular}
  }
\end{table}

\noindent \textbf{Type of padding modes.} The padding operation is to supplement pillars to the tail in neighboring maps. 
Different modes can decide what noises (extra information) would be injected to the pillars on the margin of images. 
We test four popular padding modes in Table \ref{tab-A1-2PM}. 
Among them, \textit{Replicated Padding} can inject repeated features to the marginal pillars, which might bring redundancy. 
Both the \textit{Circular} and \textit{Reflect} modes add long-distance information to those pillars, which is obviously detrimental to the locality bias. 
\textit{Zero Padding} is clean and thus achieves the best accuracy.

\begin{table}[h]
  \caption{The model accuracy (\%) on four different padding modes in the Pillars-Shift process.}
  \label{tab-A1-2PM}
  \centering
  \renewcommand\tabcolsep{4.0pt}
  \scalebox{0.9}{
  \begin{tabular}{l|cccc}
    \toprule
    Padding Mode ($p_m$)     & MIN    & C10      & C100    & Fashion \\
    \midrule
    Zero              & \textbf{78.16}      & \textbf{97.10}      & \textbf{83.86}      & \textbf{95.72}   \\
    Replicated        & \underline{78.13}    & 96.88    & 83.35    & 95.41 \\
    Circular          & 78.08    & 96.81    & \underline{83.40}    & \underline{95.45} \\
    Reflect           & 77.76    & \underline{96.89}    & 83.33    & 95.40 \\
    \bottomrule
  \end{tabular}
  }
\end{table}

% \subsubsection{Pillars-Concatenation of SPC}
\noindent \textbf{4.3.2 Pillars-Concatenation of SPC}

The Pillars-Concatenation process involves three key operations: (1) \textit{Reduce}, which enables diverse representation by transforming neighboring maps in multiple representation spaces; (2) \textit{Concatenation (Concat)}, which integrates four neighboring maps to combine local features for all pillars in parallel; and (3) \textit{Fuse}, which selectively learns and weights neighboring features to enhance the representation capabilities.
We ablate five combinations of these operations in Table \ref{tab-A2-2FusionWay} and observe that \textit{Reduce+Concat+Fuse} and \textit{Concat+Fuse} perform better than the other options. Among them, \textit{Reduce+Concat+Fuse} achieves a better trade-off between computational costs and accuracy. 

\begin{table}[h]
  \caption{Results (\%) of different ways to mix local neighbors.}
  \label{tab-A2-2FusionWay}
  \centering
  \renewcommand\tabcolsep{3.0pt}
  \scalebox{0.9}{
  \begin{tabular}{l|cccc|cc}
    \toprule
    Mixing ways       & MIN    & C10      & C100    & Fashion    & Params    & FLOPs \\
    \midrule
    Reduce+Concat+Fuse     & \underline{78.16}      & \textbf{97.10}      & \underline{83.86}      & \textbf{95.72}     & 28.4M     & 1.9G \\
    Reduce+Concat          & 77.72    & 97.02    & 83.26    & \textbf{95.72}     & 25.9M     & 1.8G \\
    Concat+Fuse          & \textbf{78.37}    & \underline{97.06}    & \textbf{83.94}    & 95.45     & 33.3M     & 2.3G \\
    Sum+Fuse          & 76.14    & 96.16    & 80.88    & 95.28     & 25.9M     & 1.8G \\
    Sum               & 74.81    & 95.33    & 75.48    & 95.34     & 23.4M     & 1.6G \\
    \bottomrule
  \end{tabular}
  }
\end{table}

% \subsubsection{Local-Global Combination Strategies}
\noindent \textbf{4.3.3 Local-Global Combination Strategies}

In Section \ref{Section 4.1} and \ref{Section 4.2}, the sMLPNet, Strip-MLP, and Caterpillar all attained excellent performance on both small- and large-scale image recognition tasks. 
This success could be attributed to the common strategy that they clearly separate the local- and global-mixing operations in their token-mixing modules.
However, both the sMLPNet and Strip-MLP missed the experiments to further explore the effects of local-global combination ways.
To fill this gap, we conduct this ablation and perform various strategies as depicted in Figure \ref{l-gways}.
We evaluate six strategies, denoted as (a), (b), (c) for sequential regimes, and (e), (f), (g) for parallel methods, by rearranging the SPC and sMLP modules in Caterpillar blocks. 
Table \ref{tab-A3-1CombineWay} shows that the simply sequential methods generally outperform the complicated parallel strategies. 
We attribute this phenomenon to the idea that the ‘High-Cohesion and Low-Coupling' principle leads to higher performance, since the internal SPC and sMLP modules are both working in sophisticated parallel ways.
Furthermore, the L-G strategy achieves the best performance.

\begin{figure}[h]
  \centering
   %\includegraphics[width=0.8\linewidth]{egfigure.eps}
   \includegraphics[width=1.0 \linewidth]{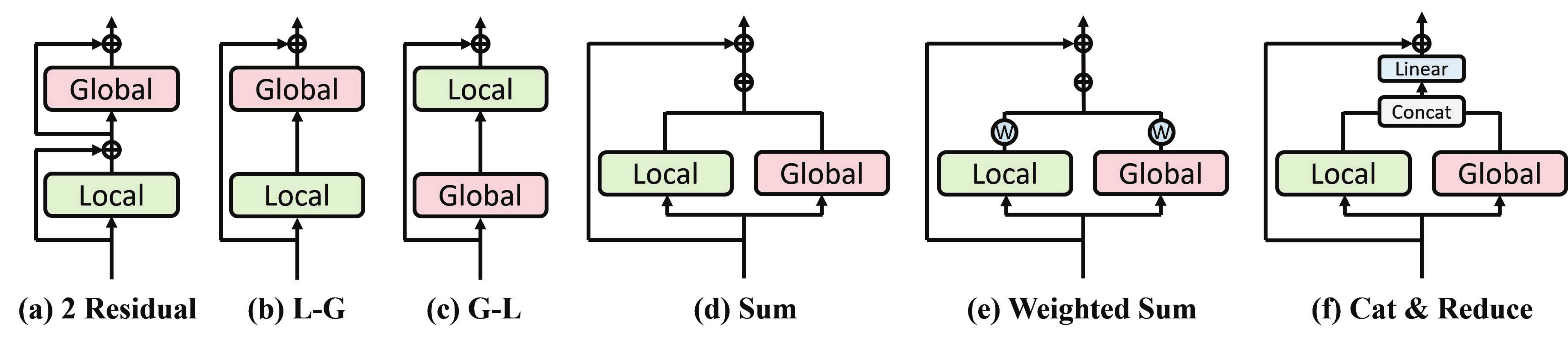}

   \caption{Different ways to combine local and global information.}
   \label{l-gways}
   \vspace{-1em}
\end{figure}

\begin{table}[h]
  \caption{Comparison (\%) between six different strategies for combining local and global information.}
  \label{tab-A3-1CombineWay}
  \centering
  \renewcommand\tabcolsep{3.0pt}
  \scalebox{0.9}{
  \begin{tabular}{l|cccc|cc}
    \toprule
    Combine ways       & MIN    & C10      & C100    & Fashion    & Params    & FLOPs \\
    \midrule
    2 Residual     & 77.06    & \underline{96.92}    & 82.51    & 95.64     & 28.4M     & 1.9G \\
    L-G (default)  & \textbf{78.16}    & \textbf{97.10}    & \textbf{83.86}     & \textbf{95.72}  & 28.4M     & 1.9G \\
    G-L          & \underline{78.09}    & 96.88    & \underline{83.45}    & \underline{95.65}     & 28.4M     & 1.9G \\
    Sum         & 76.91    & 96.70    & 82.13    & 95.53     & 28.4M     & 1.9G \\
    Weighted Sum               & 77.94    & 96.82    & 82.56    & 95.60     & 30.3M     & 2.0G \\
    Concat+Reduce           & 76.77    & 96.18    & 82.15    & 95.49     & 33.4M     & 2.3G \\
    \bottomrule
  \end{tabular}
  }
\end{table}

\subsection{Visualization}
\label{Section 4.4}

To understand how the SPC module processes image data, we visualize the feature maps encoded by SPC coupled with two control ways. 
Specifically, we build three Caterpillar-T models with the local modules of identity, convolution and SPC, and implement them on the CIFAR-100 dataset. 
Figure \ref{visualization} (in Appendix \ref{sup.1}) illustrates the feature maps of six samples, each of which is presented with 3 rows and 4 columns, where rows represent different local-mixing ways and columns are feature maps of different phases in models.
For these samples, with the \textit{(a) cattle} as an example, the patterns in SPC features are closer to the convolution and different from the identity. Since convolutional layers capably capture local features, the SPC is also capable of aggregating local information.
Furthermore, compared to convolution, the objects in SPC maps exhibit more prominent edge features and are closer to the original input image, indicating that the proposed SPC module can encode local information more elaborately and avoid redundancy issues.

\subsection{Analysis with Transfer Learning}
\label{Section 4.5}

\noindent \textbf{4.5.1 Transfer Learning Performance of Caterpillar}

In this subsection, we compare the Transfer Learning capability of the proposed Caterpillar architecture with recent SOTA models. 
Following the recent MLP-based models \cite{hou2022vip, wang2022dynamixer}, we pre-train the Caterpillar-T on the ImageNet-1k and then fine-tune it on CIFAR-10 and CIFAR-100. From Table \ref{tab-A8-2PM}, the Caterpillar attains higher scores than other representative networks with similar computational costs, indicating that Caterpillar can work well on Transfer tasks.\\

\begin{table}[h]
  \caption{The transfer learning results of models pre-trained on ImageNet-1k and fine-tuned to CIFAR-10 and CIFAR-100.}
  \label{tab-A8-2PM}
  \centering
  \renewcommand\tabcolsep{4.0pt}
  \scalebox{0.8}{
  \begin{tabular}{l|c|c|c}
    \toprule
    Networks       & Datasets    & Params  & Top-1(\%)    \\
    \midrule
    ViT-S/16 \cite{dosovitskiy2020vit}       &    & 49M     & 97.1   \\
    % ResMLP-T \cite{touvron2022resmlp}       &     & 30M     & 98.1   \\
    ViP-S/7 \cite{hou2022vip}        & CIFAR-10    & 25M     & 98.0   \\
    DynaMixer-S \cite{wang2022dynamixer}    &     & 26M     & \underline{98.2}   \\
    Caterpillar-T  &     & 29M     & \textbf{98.3}   \\
    \midrule
    ViT-S/16 \cite{dosovitskiy2020vit}       &    & 49M     & 87.1   \\
    % ResMLP-T \cite{touvron2022resmlp}       &    & 30M     & 87.0   \\
    ViP-S/7 \cite{hou2022vip}        & CIFAR-100   & 25M     & 88.4   \\
    DynaMixer-S \cite{wang2022dynamixer}    &    & 26M     & \underline{88.6}   \\
    Caterpillar-T  &    & 29M     & \textbf{89.3}   \\
    \bottomrule
  \end{tabular}
  }
\end{table}

\noindent \textbf{4.5.2 Comparison between Direct and Transfer Strategies}

Despite the remarkable Transfer capability, we highlight that Caterpillar can achieve exceptional performance on small-scale images using the 'Direct Training' (Direct) strategy (Section \ref{Section 4.1}). 
To further illustrate Caterpillar's effectiveness in data-hungry domains without relying on pre-training data, which always faces challenges to domain-shift and task-compatibility, we conduct this study.

We adopt the same datasets as in Section \ref{Section 4.1}, \textit{i.e.,} MIN, C10, C100 and Fashion, while including two more datasets in certain scientific fields of remote sensing, \textit{i.e.,} Resisc45 (R45) \cite{cheng2017remote} with 27,000 training and 4,500 testing images in 45 categories, and disease diagnosis \cite{kermany2018identifying} \textit{i.e.,} Chest\_Xray (Chest) with 5,216 training images and 624 testing images belonging to 2 classes. 
All images are resized to 224 $\times$ 224.
Then, we utilize two Caterpillar-T models as the base architectures. 
The model of ‘Transfer’ is pre-trained on the ImageNet-1K and then fine-tune on the target datasets, while the other one for ‘Direct’ is trained from scratch.
From Table \ref{tab-Sup-B}, the `Transfer’ strategy performs better on MIN, C10, and C100, while the `Direct’ strategy achieves higher scores on Fashion, R45, and Chest, which have dissimilar distributions to the pre-trained data (ImageNet-1K).
Therefore, for the data-hungry scientific tasks (especially those without the same distribution to the pre-trained natural images), directly training the deep model can be a more cost-effective approach than the `Transfer' strategy, with Caterpillar serving as the backbone architecture.

\begin{table}[h]
  \caption{Comparison (\%) between ‘Transfer Learning’ and ‘Direct Training’ strategies on six small-scale datasets. 
  % The Epochs consists two parts: for Transfer, it means pre-train + fine-tune epochs; for Direct, which has no pre-traing process, it means the whole training epochs.
  }
  \label{tab-Sup-B}
  \centering
  \renewcommand\tabcolsep{1.5pt}
  \scalebox{0.9}{
  \begin{tabular}{l|c|c|cccccc}
    \toprule
    Networks                       & Strategy    & Epochs      & MIN                & C10     & C100    & Fashion   & R45      & Chest     \\
    \midrule
    \multirow{2}*{Caterpillar-T}   & Transfer    & 300$+$30     & \textbf{95.14}   & \textbf{98.31}   & \textbf{89.30}    & 95.57    & 97.27    & 93.97    \\
    ~                              & Direct      & 0$+$300      & 86.98	          & 97.60            & 84.67	     & \textbf{96.13}        & \textbf{97.35}    & \textbf{94.29}    \\
    \bottomrule
  \end{tabular}
  }
\end{table}

\subsection{Exploration for SPC}
\label{Section 4.6}

Previous comparison between Caterpillars and sMLPNets demonstrates the potential of SPC as an alternative to convolution in plug-and-play ways (Table \ref{tab-result-M1}, \ref{tab-M3}). We further explore the SPC to serve as the main module for neural architectures. 

\noindent \textbf{Datasets.} We utilize the same large-scale benchmark of ImageNet-1K as well as the small-scale datasets of MIN, C10, C100 and Fashion, as those in Section \ref{Section 4.1} and \ref{Section 4.2}.

\begin{table}[h]
  \caption{Results (\%) of Res-18 and Res-18(SPC) on ImageNet-1K. 
 $N_{C}$ is the channel number of hidden layers in first stage.
  }
  \label{tab-A4-1}
  \centering
  \renewcommand\tabcolsep{4.0pt}
  \scalebox{0.9}{
  \begin{tabular}{l|c|c|c|c}
    \toprule
    Networks      & $N_{C}$      & Params      & FLOPs     & Top-1 (\%) \\
    \midrule
    Res-18\cite{wightman2021resnetstrike}                 & 64    & 12M    & 1.8G    & 70.6   \\
    \midrule
    \multirow{3}*{Res-18(SPC)}   & 64    & 3M    & 0.6G    & 69.1   \\
                                       & 96    & 7M    & 1.3G    & \underline{73.6}   \\
    ~                                  & 128    & 11M    & 2.2G    & \textbf{75.3}   \\
    \bottomrule
  \end{tabular}
  }
\end{table}

\begin{table}[t]
  \caption{Results (\%) of Res-18 and Res-18(SPC) on four small-scale datasets}
  \label{tab-A4-2}
  \centering
  \renewcommand\tabcolsep{2.0pt}
  \scalebox{0.9}{
  \begin{tabular}{l|c|cccc|cc}
    \toprule
    Networks      & \textit{$N_{C}$}      & MIN    & C10     & C100    & Fashion   & Params      & FLOPs     \\
    \midrule
    Res-18\cite{wightman2021resnetstrike}                & 64   & 70.95    & 95.54   & 77.66   & 95.11   & 11.2M    & 0.7G    \\
    \midrule
    \multirow{3}*{Res-18(SPC)}   & 64   & 70.10	    & 94.52     & 76.19	    & 94.90   & 2.6M    & 0.2G    \\
                                 & 96  & \underline{71.88}    & \underline{95.72}     & \underline{78.35}    & \underline{95.33}   & 5.7M    & 0.4G    \\
    ~                            & 128  & \textbf{73.24}     & \textbf{95.84}    & \textbf{79.77}  & \textbf{95.54}    & 10.2M    & 0.8G    \\
    \bottomrule
  \end{tabular}
  }
\end{table}

\noindent \textbf{Experimental Settings.} 
We adopt classic ResNet-18 (Res-18) \cite{he2016resnet} as the baseline CNN. 
Then, we replace all convolutional layers within Res-18's basic blocks with the SPC module and obtain three SPC-based variants referred to as ‘Res-18(SPC)’, with $N_{C}$ utilized as channel numbers to adjust model complexity. 
For training these models, we follow the ‘Procedure A2’ in \cite{wightman2021resnetstrike}.

\noindent \textbf{Results.} Table \ref{tab-A4-1} displays the ImageNet-1K classification results for the original Res-18 and Res-18(SPC) variants. 
As we can see, the proposed SPC module can provide higher performance than convolution with only half of the parameters (Res-18(SPC), $N_{C}$=96). 
Increasing $N_{C}$ to 128, the Res-18(SPC) reaches similar computational costs to the baseline Res-18 while achieving 4.7\% higher accuracy. 
Similar trends can be observed on small-scale recognition tasks, as shown in Table \ref{tab-A4-2}.
% Similar trends can be observed on small-scale recognition tasks, as shown in Supplementary \ref{sup:Res-18(SPC)_supp}.
This confirms that the SPC module can also be used as the main component to construct neural networks, potentially serving as an alternative to convolutional layers in independent manners.\\

\section{Conclusion}
\label{sec:conclusion}

This paper proposes the SPC module that conducts the Pillars-Shift and Pillars-Concatenation to achieve an elaborate and parallelizable aggregation of local information, with superior classification performance than convolutional layers.
Based on SPC, we introduce Caterpillar, a pure-MLP network that attains impressive scores on both small- and large-scale image recognition tasks.

% and the principle of "control variable" 
The philosophy of "simple and effective" and the principle of "control variable" have run through this work. 
Therefore, Caterpillar only replaces the DWConv with SPC module in sMLPNet and thus has more parameters. 
% We anticipate that integrating the SPC module with lightweight techniques, like depth-wise and sparse settings, will reduce computational costs and further improve module performance.
We anticipate that integrating the SPC module with lightweight techniques, like depth-wise, will further reduce computational costs.
% In the future, we will improve the Caterpillar and SPC into more lightweight versions.
Additionally, the experiments are primarily conducted on the most fundamental classification tasks, since SPC and Caterpillar are introduced for the first time.
% the most fundamental and representative classification task.
% for a fair and uniform comparison to recent SOTA methods. 
We hope the SPC and Caterpillar can be explored in broader tasks like detection and segmentation, particularly in data-hungry domains.

\section{Acknowledgments}

This work was supported by the National Key Research and Development Program of China under Grant (No. 2022YFA1004100), and the National Natural Science Foundation of China (No. 62276052).

% Guangxi Natural Science Foundation 2023GXNSFBA026010, Research Fund of Guangxi Key Lab of Multi-source Information Mining Security 22-A-03-02, Guangxi “Bagui” Teams for Innovation and Research, China, and the National Natural Science Foundation of China No. 62166003.

% \section{Appendices}

% If your work needs an appendix, add it before the
% ``\verb|\end{document}|'' command at the conclusion of your source
% document.

% Start the appendix with the ``\verb|appendix|'' command:
% \begin{verbatim}
%   \appendix
% \end{verbatim}
% and note that in the appendix, sections are lettered, not
% numbered. This document has two appendices, demonstrating the section
% and subsection identification method.

\bibliographystyle{ACM-Reference-Format}
\balance
\bibliography{sample-base}


\begin{thebibliography}{61}

%%% the Web address.

\ifx \showCODEN    \undefined \def \showCODEN     #1{\unskip}     \fi
\ifx \showDOI      \undefined \def \showDOI       #1{#1}\fi
\ifx \showISBNx    \undefined \def \showISBNx     #1{\unskip}     \fi
\ifx \showISBNxiii \undefined \def \showISBNxiii  #1{\unskip}     \fi
\ifx \showISSN     \undefined \def \showISSN      #1{\unskip}     \fi
\ifx \showLCCN     \undefined \def \showLCCN      #1{\unskip}     \fi
\ifx \shownote     \undefined \def \shownote      #1{#1}          \fi
\ifx \showarticletitle \undefined \def \showarticletitle #1{#1}   \fi
\ifx \showURL      \undefined \def \showURL       {\relax}        \fi
% invisible to TeX
\providecommand\bibfield[2]{#2}
\providecommand\bibinfo[2]{#2}
\providecommand\natexlab[1]{#1}
\providecommand\showeprint[2][]{arXiv:#2}

\bibitem[Cao et~al\mbox{.}(2023)]%
        {cao2023stripmlp}
\bibfield{author}{\bibinfo{person}{Guiping Cao}, \bibinfo{person}{Shengda Luo}, \bibinfo{person}{Wenjian Huang}, \bibinfo{person}{Xiangyuan Lan}, \bibinfo{person}{Dongmei Jiang}, \bibinfo{person}{Yaowei Wang}, {and} \bibinfo{person}{Jianguo Zhang}.} \bibinfo{year}{2023}\natexlab{}.
\newblock \showarticletitle{Strip-MLP: Efficient Token Interaction for Vision MLP}. In \bibinfo{booktitle}{\emph{ICCV}}. \bibinfo{pages}{1494--1504}.
\newblock


\bibitem[Chen et~al\mbox{.}(2023)]%
        {chen2023cyclemlp}
\bibfield{author}{\bibinfo{person}{Shoufa Chen}, \bibinfo{person}{Enze Xie}, \bibinfo{person}{Chongjian Ge}, \bibinfo{person}{Runjian Chen}, \bibinfo{person}{Ding Liang}, {and} \bibinfo{person}{Ping Luo}.} \bibinfo{year}{2023}\natexlab{}.
\newblock \showarticletitle{CycleMLP: A MLP-like Architecture for Dense Visual Predictions}.
\newblock \bibinfo{journal}{\emph{IEEE TPAMI}} (\bibinfo{year}{2023}).
\newblock


\bibitem[Chen et~al\mbox{.}(2019)]%
        {chen2019sparseshift}
\bibfield{author}{\bibinfo{person}{Weijie Chen}, \bibinfo{person}{Di Xie}, \bibinfo{person}{Yuan Zhang}, {and} \bibinfo{person}{Shiliang Pu}.} \bibinfo{year}{2019}\natexlab{}.
\newblock \showarticletitle{All you need is a few shifts: Designing efficient convolutional neural networks for image classification}. In \bibinfo{booktitle}{\emph{CVPR}}. \bibinfo{pages}{7241--7250}.
\newblock


\bibitem[Chen et~al\mbox{.}(2021)]%
        {chen2021x}
\bibfield{author}{\bibinfo{person}{Xuanhong Chen}, \bibinfo{person}{Hang Wang}, {and} \bibinfo{person}{Bingbing Ni}.} \bibinfo{year}{2021}\natexlab{}.
\newblock \showarticletitle{X-volution: On the unification of convolution and self-attention}.
\newblock \bibinfo{journal}{\emph{arXiv preprint arXiv:2106.02253}} (\bibinfo{year}{2021}).
\newblock


\bibitem[Cheng et~al\mbox{.}(2017)]%
        {cheng2017remote}
\bibfield{author}{\bibinfo{person}{Gong Cheng}, \bibinfo{person}{Junwei Han}, {and} \bibinfo{person}{Xiaoqiang Lu}.} \bibinfo{year}{2017}\natexlab{}.
\newblock \showarticletitle{Remote sensing image scene classification: Benchmark and state of the art}.
\newblock \bibinfo{journal}{\emph{Proc. IEEE}} \bibinfo{volume}{105}, \bibinfo{number}{10} (\bibinfo{year}{2017}), \bibinfo{pages}{1865--1883}.
\newblock


\bibitem[Cubuk et~al\mbox{.}(2020)]%
        {cubuk2020randaugment}
\bibfield{author}{\bibinfo{person}{Ekin~D Cubuk}, \bibinfo{person}{Barret Zoph}, \bibinfo{person}{Jonathon Shlens}, {and} \bibinfo{person}{Quoc~V Le}.} \bibinfo{year}{2020}\natexlab{}.
\newblock \showarticletitle{Randaugment: Practical automated data augmentation with a reduced search space}. In \bibinfo{booktitle}{\emph{CVPR}}. \bibinfo{pages}{702--703}.
\newblock


\bibitem[Dai et~al\mbox{.}(2017)]%
        {dai2017deformableConv}
\bibfield{author}{\bibinfo{person}{Jifeng Dai}, \bibinfo{person}{Haozhi Qi}, \bibinfo{person}{Yuwen Xiong}, \bibinfo{person}{Yi Li}, \bibinfo{person}{Guodong Zhang}, \bibinfo{person}{Han Hu}, {and} \bibinfo{person}{Yichen Wei}.} \bibinfo{year}{2017}\natexlab{}.
\newblock \showarticletitle{Deformable convolutional networks}. In \bibinfo{booktitle}{\emph{ICCV}}. \bibinfo{pages}{764--773}.
\newblock


\bibitem[Deng et~al\mbox{.}(2009)]%
        {deng2009imagenet1k}
\bibfield{author}{\bibinfo{person}{Jia Deng}, \bibinfo{person}{Wei Dong}, \bibinfo{person}{Richard Socher}, \bibinfo{person}{Li-Jia Li}, \bibinfo{person}{Kai Li}, {and} \bibinfo{person}{Li Fei-Fei}.} \bibinfo{year}{2009}\natexlab{}.
\newblock \showarticletitle{Imagenet: A large-scale hierarchical image database}. In \bibinfo{booktitle}{\emph{CVPR}}. \bibinfo{pages}{248--255}.
\newblock


\bibitem[Dosovitskiy et~al\mbox{.}(2021)]%
        {dosovitskiy2020vit}
\bibfield{author}{\bibinfo{person}{Alexey Dosovitskiy}, \bibinfo{person}{Lucas Beyer}, \bibinfo{person}{Alexander Kolesnikov}, \bibinfo{person}{Dirk Weissenborn}, \bibinfo{person}{Xiaohua Zhai}, \bibinfo{person}{Thomas Unterthiner}, \bibinfo{person}{Mostafa Dehghani}, \bibinfo{person}{Matthias Minderer}, \bibinfo{person}{Georg Heigold}, \bibinfo{person}{Sylvain Gelly}, {et~al\mbox{.}}} \bibinfo{year}{2021}\natexlab{}.
\newblock \showarticletitle{An image is worth 16x16 words: Transformers for image recognition at scale}.
\newblock \bibinfo{journal}{\emph{ICLR}} (\bibinfo{year}{2021}).
\newblock


\bibitem[Fukushima(1975)]%
        {fukushima1975cognitron}
\bibfield{author}{\bibinfo{person}{Kunihiko Fukushima}.} \bibinfo{year}{1975}\natexlab{}.
\newblock \showarticletitle{Cognitron: A self-organizing multilayered neural network}.
\newblock \bibinfo{journal}{\emph{Biological cybernetics}} \bibinfo{volume}{20}, \bibinfo{number}{3-4} (\bibinfo{year}{1975}), \bibinfo{pages}{121--136}.
\newblock


\bibitem[Fukushima(1980)]%
        {fukushima1980neocognitron}
\bibfield{author}{\bibinfo{person}{Kunihiko Fukushima}.} \bibinfo{year}{1980}\natexlab{}.
\newblock \showarticletitle{Neocognitron: A self-organizing neural network model for a mechanism of pattern recognition unaffected by shift in position}.
\newblock \bibinfo{journal}{\emph{Biological cybernetics}} \bibinfo{volume}{36}, \bibinfo{number}{4} (\bibinfo{year}{1980}), \bibinfo{pages}{193--202}.
\newblock


\bibitem[Guo et~al\mbox{.}(2022)]%
        {guo2022hiremlp}
\bibfield{author}{\bibinfo{person}{Jianyuan Guo}, \bibinfo{person}{Yehui Tang}, \bibinfo{person}{Kai Han}, \bibinfo{person}{Xinghao Chen}, \bibinfo{person}{Han Wu}, \bibinfo{person}{Chao Xu}, \bibinfo{person}{Chang Xu}, {and} \bibinfo{person}{Yunhe Wang}.} \bibinfo{year}{2022}\natexlab{}.
\newblock \showarticletitle{Hire-mlp: Vision mlp via hierarchical rearrangement}. In \bibinfo{booktitle}{\emph{CVPR}}. \bibinfo{pages}{826--836}.
\newblock


\bibitem[Guo et~al\mbox{.}(2023)]%
        {guo2023visual}
\bibfield{author}{\bibinfo{person}{Meng-Hao Guo}, \bibinfo{person}{Cheng-Ze Lu}, \bibinfo{person}{Zheng-Ning Liu}, \bibinfo{person}{Ming-Ming Cheng}, {and} \bibinfo{person}{Shi-Min Hu}.} \bibinfo{year}{2023}\natexlab{}.
\newblock \showarticletitle{Visual attention network}.
\newblock \bibinfo{journal}{\emph{Computational Visual Media}} \bibinfo{volume}{9}, \bibinfo{number}{4} (\bibinfo{year}{2023}), \bibinfo{pages}{733--752}.
\newblock


\bibitem[Hassani et~al\mbox{.}(2021)]%
        {hassani2021cct}
\bibfield{author}{\bibinfo{person}{Ali Hassani}, \bibinfo{person}{Steven Walton}, \bibinfo{person}{Nikhil Shah}, \bibinfo{person}{Abulikemu Abuduweili}, \bibinfo{person}{Jiachen Li}, {and} \bibinfo{person}{Humphrey Shi}.} \bibinfo{year}{2021}\natexlab{}.
\newblock \showarticletitle{Escaping the big data paradigm with compact transformers}.
\newblock \bibinfo{journal}{\emph{arXiv preprint arXiv:2104.05704}} (\bibinfo{year}{2021}).
\newblock


\bibitem[He et~al\mbox{.}(2022)]%
        {he2022mae}
\bibfield{author}{\bibinfo{person}{Kaiming He}, \bibinfo{person}{Xinlei Chen}, \bibinfo{person}{Saining Xie}, \bibinfo{person}{Yanghao Li}, \bibinfo{person}{Piotr Doll{\'a}r}, {and} \bibinfo{person}{Ross Girshick}.} \bibinfo{year}{2022}\natexlab{}.
\newblock \showarticletitle{Masked autoencoders are scalable vision learners}. In \bibinfo{booktitle}{\emph{CVPR}}. \bibinfo{pages}{16000--16009}.
\newblock


\bibitem[He et~al\mbox{.}(2016)]%
        {he2016resnet}
\bibfield{author}{\bibinfo{person}{Kaiming He}, \bibinfo{person}{Xiangyu Zhang}, \bibinfo{person}{Shaoqing Ren}, {and} \bibinfo{person}{Jian Sun}.} \bibinfo{year}{2016}\natexlab{}.
\newblock \showarticletitle{Deep residual learning for image recognition}. In \bibinfo{booktitle}{\emph{CVPR}}. \bibinfo{pages}{770--778}.
\newblock


\bibitem[Hou et~al\mbox{.}(2022)]%
        {hou2022vip}
\bibfield{author}{\bibinfo{person}{Qibin Hou}, \bibinfo{person}{Zihang Jiang}, \bibinfo{person}{Li Yuan}, \bibinfo{person}{Ming-Ming Cheng}, \bibinfo{person}{Shuicheng Yan}, {and} \bibinfo{person}{Jiashi Feng}.} \bibinfo{year}{2022}\natexlab{}.
\newblock \showarticletitle{Vision permutator: A permutable mlp-like architecture for visual recognition}.
\newblock \bibinfo{journal}{\emph{IEEE TPAMI}} \bibinfo{volume}{45}, \bibinfo{number}{1} (\bibinfo{year}{2022}), \bibinfo{pages}{1328--1334}.
\newblock


\bibitem[Hubel and Wiesel(1962)]%
        {hubel1962receptive}
\bibfield{author}{\bibinfo{person}{David~H Hubel} {and} \bibinfo{person}{Torsten~N Wiesel}.} \bibinfo{year}{1962}\natexlab{}.
\newblock \showarticletitle{Receptive fields, binocular interaction and functional architecture in the cat's visual cortex}.
\newblock \bibinfo{journal}{\emph{The Journal of physiology}} \bibinfo{volume}{160}, \bibinfo{number}{1} (\bibinfo{year}{1962}), \bibinfo{pages}{106}.
\newblock


\bibitem[Hubel and Wiesel(1965)]%
        {hubel1965receptive}
\bibfield{author}{\bibinfo{person}{David~H Hubel} {and} \bibinfo{person}{Torsten~N Wiesel}.} \bibinfo{year}{1965}\natexlab{}.
\newblock \showarticletitle{Receptive fields and functional architecture in two nonstriate visual areas (18 and 19) of the cat}.
\newblock \bibinfo{journal}{\emph{Journal of neurophysiology}} \bibinfo{volume}{28}, \bibinfo{number}{2} (\bibinfo{year}{1965}), \bibinfo{pages}{229--289}.
\newblock


\bibitem[Kermany et~al\mbox{.}(2018)]%
        {kermany2018identifying}
\bibfield{author}{\bibinfo{person}{Daniel~S Kermany}, \bibinfo{person}{Michael Goldbaum}, \bibinfo{person}{Wenjia Cai}, \bibinfo{person}{Carolina~CS Valentim}, \bibinfo{person}{Huiying Liang}, \bibinfo{person}{Sally~L Baxter}, \bibinfo{person}{Alex McKeown}, \bibinfo{person}{Ge Yang}, \bibinfo{person}{Xiaokang Wu}, \bibinfo{person}{Fangbing Yan}, {et~al\mbox{.}}} \bibinfo{year}{2018}\natexlab{}.
\newblock \showarticletitle{Identifying medical diagnoses and treatable diseases by image-based deep learning}.
\newblock \bibinfo{journal}{\emph{cell}} \bibinfo{volume}{172}, \bibinfo{number}{5} (\bibinfo{year}{2018}), \bibinfo{pages}{1122--1131}.
\newblock


\bibitem[Krizhevsky et~al\mbox{.}(2009)]%
        {krizhevsky2009cifar}
\bibfield{author}{\bibinfo{person}{Alex Krizhevsky}, \bibinfo{person}{Geoffrey Hinton}, {et~al\mbox{.}}} \bibinfo{year}{2009}\natexlab{}.
\newblock \showarticletitle{Learning multiple layers of features from tiny images}.
\newblock \bibinfo{journal}{\emph{Citeseer, Tech. Rep.}} (\bibinfo{year}{2009}).
\newblock


\bibitem[Krizhevsky et~al\mbox{.}(2017)]%
        {krizhevsky2017alexnet}
\bibfield{author}{\bibinfo{person}{Alex Krizhevsky}, \bibinfo{person}{Ilya Sutskever}, {and} \bibinfo{person}{Geoffrey~E Hinton}.} \bibinfo{year}{2017}\natexlab{}.
\newblock \showarticletitle{Imagenet classification with deep convolutional neural networks}.
\newblock \bibinfo{journal}{\emph{Commun. ACM}} \bibinfo{volume}{60}, \bibinfo{number}{6} (\bibinfo{year}{2017}), \bibinfo{pages}{84--90}.
\newblock


\bibitem[LeCun et~al\mbox{.}(1989b)]%
        {lecun1989conv1}
\bibfield{author}{\bibinfo{person}{Yann LeCun} {et~al\mbox{.}}} \bibinfo{year}{1989}\natexlab{b}.
\newblock \showarticletitle{Generalization and network design strategies}.
\newblock \bibinfo{journal}{\emph{Connectionism in perspective}} \bibinfo{volume}{19}, \bibinfo{number}{143-155} (\bibinfo{year}{1989}), \bibinfo{pages}{18}.
\newblock


\bibitem[LeCun et~al\mbox{.}(1989a)]%
        {lecun1989conv2}
\bibfield{author}{\bibinfo{person}{Yann LeCun}, \bibinfo{person}{Bernhard Boser}, \bibinfo{person}{John~S Denker}, \bibinfo{person}{Donnie Henderson}, \bibinfo{person}{Richard~E Howard}, \bibinfo{person}{Wayne Hubbard}, {and} \bibinfo{person}{Lawrence~D Jackel}.} \bibinfo{year}{1989}\natexlab{a}.
\newblock \showarticletitle{Backpropagation applied to handwritten zip code recognition}.
\newblock \bibinfo{journal}{\emph{Neural computation}} \bibinfo{volume}{1}, \bibinfo{number}{4} (\bibinfo{year}{1989}), \bibinfo{pages}{541--551}.
\newblock


\bibitem[LeCun et~al\mbox{.}(1998)]%
        {lecun1998lenet}
\bibfield{author}{\bibinfo{person}{Yann LeCun}, \bibinfo{person}{L{\'e}on Bottou}, \bibinfo{person}{Yoshua Bengio}, {and} \bibinfo{person}{Patrick Haffner}.} \bibinfo{year}{1998}\natexlab{}.
\newblock \showarticletitle{Gradient-based learning applied to document recognition}.
\newblock \bibinfo{journal}{\emph{Proc. IEEE}} \bibinfo{volume}{86}, \bibinfo{number}{11} (\bibinfo{year}{1998}), \bibinfo{pages}{2278--2324}.
\newblock


\bibitem[Li et~al\mbox{.}(2023b)]%
        {li2023simple}
\bibfield{author}{\bibinfo{person}{Dasong Li}, \bibinfo{person}{Xiaoyu Shi}, \bibinfo{person}{Yi Zhang}, \bibinfo{person}{Ka~Chun Cheung}, \bibinfo{person}{Simon See}, \bibinfo{person}{Xiaogang Wang}, \bibinfo{person}{Hongwei Qin}, {and} \bibinfo{person}{Hongsheng Li}.} \bibinfo{year}{2023}\natexlab{b}.
\newblock \showarticletitle{A simple baseline for video restoration with grouped spatial-temporal shift}. In \bibinfo{booktitle}{\emph{CVPR}}. \bibinfo{pages}{9822--9832}.
\newblock


\bibitem[Li et~al\mbox{.}(2023a)]%
        {li2023convmlp}
\bibfield{author}{\bibinfo{person}{Jiachen Li}, \bibinfo{person}{Ali Hassani}, \bibinfo{person}{Steven Walton}, {and} \bibinfo{person}{Humphrey Shi}.} \bibinfo{year}{2023}\natexlab{a}.
\newblock \showarticletitle{Convmlp: Hierarchical convolutional mlps for vision}. In \bibinfo{booktitle}{\emph{CVPR}}. \bibinfo{pages}{6306--6315}.
\newblock


\bibitem[Li et~al\mbox{.}(2023c)]%
        {li2023moganet}
\bibfield{author}{\bibinfo{person}{Siyuan Li}, \bibinfo{person}{Zedong Wang}, \bibinfo{person}{Zicheng Liu}, \bibinfo{person}{Cheng Tan}, \bibinfo{person}{Haitao Lin}, \bibinfo{person}{Di Wu}, \bibinfo{person}{Zhiyuan Chen}, \bibinfo{person}{Jiangbin Zheng}, {and} \bibinfo{person}{Stan~Z Li}.} \bibinfo{year}{2023}\natexlab{c}.
\newblock \showarticletitle{Moganet: Multi-order gated aggregation network}. In \bibinfo{booktitle}{\emph{ICLR}}.
\newblock


\bibitem[Lian et~al\mbox{.}(2021)]%
        {lian2021asmlp}
\bibfield{author}{\bibinfo{person}{Dongze Lian}, \bibinfo{person}{Zehao Yu}, \bibinfo{person}{Xing Sun}, {and} \bibinfo{person}{Shenghua Gao}.} \bibinfo{year}{2021}\natexlab{}.
\newblock \showarticletitle{As-mlp: An axial shifted mlp architecture for vision}.
\newblock \bibinfo{journal}{\emph{arXiv preprint arXiv:2107.08391}} (\bibinfo{year}{2021}).
\newblock


\bibitem[Lin et~al\mbox{.}(2019)]%
        {lin2019paricalshift}
\bibfield{author}{\bibinfo{person}{Ji Lin}, \bibinfo{person}{Chuang Gan}, {and} \bibinfo{person}{Song Han}.} \bibinfo{year}{2019}\natexlab{}.
\newblock \showarticletitle{Tsm: Temporal shift module for efficient video understanding}. In \bibinfo{booktitle}{\emph{CVPR}}. \bibinfo{pages}{7083--7093}.
\newblock


\bibitem[Liu et~al\mbox{.}(2021a)]%
        {liu2021gmlp}
\bibfield{author}{\bibinfo{person}{Hanxiao Liu}, \bibinfo{person}{Zihang Dai}, \bibinfo{person}{David So}, {and} \bibinfo{person}{Quoc~V Le}.} \bibinfo{year}{2021}\natexlab{a}.
\newblock \showarticletitle{Pay attention to mlps}.
\newblock \bibinfo{journal}{\emph{NeurIPS}}  \bibinfo{volume}{34} (\bibinfo{year}{2021}), \bibinfo{pages}{9204--9215}.
\newblock


\bibitem[Liu et~al\mbox{.}(2022a)]%
        {liumore}
\bibfield{author}{\bibinfo{person}{Shiwei Liu}, \bibinfo{person}{Tianlong Chen}, \bibinfo{person}{Xiaohan Chen}, \bibinfo{person}{Xuxi Chen}, \bibinfo{person}{Qiao Xiao}, \bibinfo{person}{Boqian Wu}, \bibinfo{person}{Tommi K{\"a}rkk{\"a}inen}, \bibinfo{person}{Mykola Pechenizkiy}, \bibinfo{person}{Decebal~Constantin Mocanu}, {and} \bibinfo{person}{Zhangyang Wang}.} \bibinfo{year}{2022}\natexlab{a}.
\newblock \showarticletitle{More ConvNets in the 2020s: Scaling up Kernels Beyond 51x51 using Sparsity}. In \bibinfo{booktitle}{\emph{ICLR}}.
\newblock


\bibitem[Liu et~al\mbox{.}(2021b)]%
        {liu2021swin}
\bibfield{author}{\bibinfo{person}{Ze Liu}, \bibinfo{person}{Yutong Lin}, \bibinfo{person}{Yue Cao}, \bibinfo{person}{Han Hu}, \bibinfo{person}{Yixuan Wei}, \bibinfo{person}{Zheng Zhang}, \bibinfo{person}{Stephen Lin}, {and} \bibinfo{person}{Baining Guo}.} \bibinfo{year}{2021}\natexlab{b}.
\newblock \showarticletitle{Swin transformer: Hierarchical vision transformer using shifted windows}. In \bibinfo{booktitle}{\emph{ICCV}}. \bibinfo{pages}{10012--10022}.
\newblock


\bibitem[Liu et~al\mbox{.}(2022b)]%
        {liu2022convnet}
\bibfield{author}{\bibinfo{person}{Zhuang Liu}, \bibinfo{person}{Hanzi Mao}, \bibinfo{person}{Chao-Yuan Wu}, \bibinfo{person}{Christoph Feichtenhofer}, \bibinfo{person}{Trevor Darrell}, {and} \bibinfo{person}{Saining Xie}.} \bibinfo{year}{2022}\natexlab{b}.
\newblock \showarticletitle{A convnet for the 2020s}. In \bibinfo{booktitle}{\emph{CVPR}}. \bibinfo{pages}{11976--11986}.
\newblock


\bibitem[Loshchilov and Hutter(2019)]%
        {loshchilov2017adamw}
\bibfield{author}{\bibinfo{person}{Ilya Loshchilov} {and} \bibinfo{person}{Frank Hutter}.} \bibinfo{year}{2019}\natexlab{}.
\newblock \showarticletitle{Decoupled weight decay regularization}.
\newblock \bibinfo{journal}{\emph{ICLR}} (\bibinfo{year}{2019}).
\newblock


\bibitem[Qin et~al\mbox{.}(2018)]%
        {qin2018diagonalwise}
\bibfield{author}{\bibinfo{person}{Zheng Qin}, \bibinfo{person}{Zhaoning Zhang}, \bibinfo{person}{Dongsheng Li}, \bibinfo{person}{Yiming Zhang}, {and} \bibinfo{person}{Yuxing Peng}.} \bibinfo{year}{2018}\natexlab{}.
\newblock \showarticletitle{Diagonalwise refactorization: An efficient training method for depthwise convolutions}. In \bibinfo{booktitle}{\emph{IJCNN}}. IEEE, \bibinfo{pages}{1--8}.
\newblock


\bibitem[Radosavovic et~al\mbox{.}(2020)]%
        {radosavovic2020regnet}
\bibfield{author}{\bibinfo{person}{Ilija Radosavovic}, \bibinfo{person}{Raj~Prateek Kosaraju}, \bibinfo{person}{Ross Girshick}, \bibinfo{person}{Kaiming He}, {and} \bibinfo{person}{Piotr Doll{\'a}r}.} \bibinfo{year}{2020}\natexlab{}.
\newblock \showarticletitle{Designing network design spaces}. In \bibinfo{booktitle}{\emph{CVPR}}. \bibinfo{pages}{10428--10436}.
\newblock


\bibitem[Rao et~al\mbox{.}(2022)]%
        {rao2022hornet}
\bibfield{author}{\bibinfo{person}{Yongming Rao}, \bibinfo{person}{Wenliang Zhao}, \bibinfo{person}{Yansong Tang}, \bibinfo{person}{Jie Zhou}, \bibinfo{person}{Ser~Nam Lim}, {and} \bibinfo{person}{Jiwen Lu}.} \bibinfo{year}{2022}\natexlab{}.
\newblock \showarticletitle{Hornet: Efficient high-order spatial interactions with recursive gated convolutions}.
\newblock \bibinfo{journal}{\emph{NeurIPS}}  \bibinfo{volume}{35} (\bibinfo{year}{2022}), \bibinfo{pages}{10353--10366}.
\newblock


\bibitem[Simonyan and Zisserman(2015)]%
        {simonyan2014vggnet}
\bibfield{author}{\bibinfo{person}{Karen Simonyan} {and} \bibinfo{person}{Andrew Zisserman}.} \bibinfo{year}{2015}\natexlab{}.
\newblock \showarticletitle{Very deep convolutional networks for large-scale image recognition}.
\newblock \bibinfo{journal}{\emph{ICLR}} (\bibinfo{year}{2015}).
\newblock


\bibitem[Tang et~al\mbox{.}(2022b)]%
        {tang2022sMLPNet}
\bibfield{author}{\bibinfo{person}{Chuanxin Tang}, \bibinfo{person}{Yucheng Zhao}, \bibinfo{person}{Guangting Wang}, \bibinfo{person}{Chong Luo}, \bibinfo{person}{Wenxuan Xie}, {and} \bibinfo{person}{Wenjun Zeng}.} \bibinfo{year}{2022}\natexlab{b}.
\newblock \showarticletitle{Sparse MLP for image recognition: Is self-attention really necessary?}. In \bibinfo{booktitle}{\emph{AAAI}}, Vol.~\bibinfo{volume}{36}. \bibinfo{pages}{2344--2351}.
\newblock


\bibitem[Tang et~al\mbox{.}(2022a)]%
        {tang2022wavemlp}
\bibfield{author}{\bibinfo{person}{Yehui Tang}, \bibinfo{person}{Kai Han}, \bibinfo{person}{Jianyuan Guo}, \bibinfo{person}{Chang Xu}, \bibinfo{person}{Yanxi Li}, \bibinfo{person}{Chao Xu}, {and} \bibinfo{person}{Yunhe Wang}.} \bibinfo{year}{2022}\natexlab{a}.
\newblock \showarticletitle{An image patch is a wave: Phase-aware vision mlp}. In \bibinfo{booktitle}{\emph{CVPR}}. \bibinfo{pages}{10935--10944}.
\newblock


\bibitem[Tolstikhin et~al\mbox{.}(2021)]%
        {tolstikhin2021mlp-mixer}
\bibfield{author}{\bibinfo{person}{Ilya~O Tolstikhin}, \bibinfo{person}{Neil Houlsby}, \bibinfo{person}{Alexander Kolesnikov}, \bibinfo{person}{Lucas Beyer}, \bibinfo{person}{Xiaohua Zhai}, \bibinfo{person}{Thomas Unterthiner}, \bibinfo{person}{Jessica Yung}, \bibinfo{person}{Andreas Steiner}, \bibinfo{person}{Daniel Keysers}, \bibinfo{person}{Jakob Uszkoreit}, {et~al\mbox{.}}} \bibinfo{year}{2021}\natexlab{}.
\newblock \showarticletitle{Mlp-mixer: An all-mlp architecture for vision}.
\newblock \bibinfo{journal}{\emph{NeurIPS}}  \bibinfo{volume}{34} (\bibinfo{year}{2021}), \bibinfo{pages}{24261--24272}.
\newblock


\bibitem[Touvron et~al\mbox{.}(2022)]%
        {touvron2022resmlp}
\bibfield{author}{\bibinfo{person}{Hugo Touvron}, \bibinfo{person}{Piotr Bojanowski}, \bibinfo{person}{Mathilde Caron}, \bibinfo{person}{Matthieu Cord}, \bibinfo{person}{Alaaeldin El-Nouby}, \bibinfo{person}{Edouard Grave}, \bibinfo{person}{Gautier Izacard}, \bibinfo{person}{Armand Joulin}, \bibinfo{person}{Gabriel Synnaeve}, \bibinfo{person}{Jakob Verbeek}, {et~al\mbox{.}}} \bibinfo{year}{2022}\natexlab{}.
\newblock \showarticletitle{Resmlp: Feedforward networks for image classification with data-efficient training}.
\newblock \bibinfo{journal}{\emph{IEEE TPAMI}} (\bibinfo{year}{2022}).
\newblock


\bibitem[Touvron et~al\mbox{.}(2021)]%
        {touvron2021deit}
\bibfield{author}{\bibinfo{person}{Hugo Touvron}, \bibinfo{person}{Matthieu Cord}, \bibinfo{person}{Matthijs Douze}, \bibinfo{person}{Francisco Massa}, \bibinfo{person}{Alexandre Sablayrolles}, {and} \bibinfo{person}{Herv{\'e} J{\'e}gou}.} \bibinfo{year}{2021}\natexlab{}.
\newblock \showarticletitle{Training data-efficient image transformers \& distillation through attention}. In \bibinfo{booktitle}{\emph{ICML}}. \bibinfo{pages}{10347--10357}.
\newblock


\bibitem[Trockman and Kolter(2022)]%
        {trockman2022convmixer}
\bibfield{author}{\bibinfo{person}{Asher Trockman} {and} \bibinfo{person}{J~Zico Kolter}.} \bibinfo{year}{2022}\natexlab{}.
\newblock \showarticletitle{Patches are all you need?}
\newblock \bibinfo{journal}{\emph{arXiv preprint arXiv:2201.09792}} (\bibinfo{year}{2022}).
\newblock


\bibitem[Vaswani et~al\mbox{.}(2017)]%
        {vaswani2017transformer}
\bibfield{author}{\bibinfo{person}{Ashish Vaswani}, \bibinfo{person}{Noam Shazeer}, \bibinfo{person}{Niki Parmar}, \bibinfo{person}{Jakob Uszkoreit}, \bibinfo{person}{Llion Jones}, \bibinfo{person}{Aidan~N Gomez}, \bibinfo{person}{{\L}ukasz Kaiser}, {and} \bibinfo{person}{Illia Polosukhin}.} \bibinfo{year}{2017}\natexlab{}.
\newblock \showarticletitle{Attention is all you need}.
\newblock \bibinfo{journal}{\emph{NeurIPS}}  \bibinfo{volume}{30} (\bibinfo{year}{2017}).
\newblock


\bibitem[Vinyals et~al\mbox{.}(2016)]%
        {vinyals2016mini-imagenet}
\bibfield{author}{\bibinfo{person}{Oriol Vinyals}, \bibinfo{person}{Charles Blundell}, \bibinfo{person}{Timothy Lillicrap}, \bibinfo{person}{Daan Wierstra}, {et~al\mbox{.}}} \bibinfo{year}{2016}\natexlab{}.
\newblock \showarticletitle{Matching networks for one shot learning}.
\newblock \bibinfo{journal}{\emph{NeurIPS}}  \bibinfo{volume}{29} (\bibinfo{year}{2016}).
\newblock


\bibitem[Wang et~al\mbox{.}(2023)]%
        {wang2023internimage}
\bibfield{author}{\bibinfo{person}{Wenhai Wang}, \bibinfo{person}{Jifeng Dai}, \bibinfo{person}{Zhe Chen}, \bibinfo{person}{Zhenhang Huang}, \bibinfo{person}{Zhiqi Li}, \bibinfo{person}{Xizhou Zhu}, \bibinfo{person}{Xiaowei Hu}, \bibinfo{person}{Tong Lu}, \bibinfo{person}{Lewei Lu}, \bibinfo{person}{Hongsheng Li}, {et~al\mbox{.}}} \bibinfo{year}{2023}\natexlab{}.
\newblock \showarticletitle{Internimage: Exploring large-scale vision foundation models with deformable convolutions}. In \bibinfo{booktitle}{\emph{CVPR}}. \bibinfo{pages}{14408--14419}.
\newblock


\bibitem[Wang et~al\mbox{.}(2021)]%
        {wang2021pvt}
\bibfield{author}{\bibinfo{person}{Wenhai Wang}, \bibinfo{person}{Enze Xie}, \bibinfo{person}{Xiang Li}, \bibinfo{person}{Deng-Ping Fan}, \bibinfo{person}{Kaitao Song}, \bibinfo{person}{Ding Liang}, \bibinfo{person}{Tong Lu}, \bibinfo{person}{Ping Luo}, {and} \bibinfo{person}{Ling Shao}.} \bibinfo{year}{2021}\natexlab{}.
\newblock \showarticletitle{Pyramid vision transformer: A versatile backbone for dense prediction without convolutions}. In \bibinfo{booktitle}{\emph{ICCV}}. \bibinfo{pages}{568--578}.
\newblock


\bibitem[Wang et~al\mbox{.}(2022)]%
        {wang2022dynamixer}
\bibfield{author}{\bibinfo{person}{Ziyu Wang}, \bibinfo{person}{Wenhao Jiang}, \bibinfo{person}{Yiming~M Zhu}, \bibinfo{person}{Li Yuan}, \bibinfo{person}{Yibing Song}, {and} \bibinfo{person}{Wei Liu}.} \bibinfo{year}{2022}\natexlab{}.
\newblock \showarticletitle{Dynamixer: a vision MLP architecture with dynamic mixing}. In \bibinfo{booktitle}{\emph{ICML}}. \bibinfo{pages}{22691--22701}.
\newblock


\bibitem[Wightman et~al\mbox{.}(2021)]%
        {wightman2021resnetstrike}
\bibfield{author}{\bibinfo{person}{Ross Wightman}, \bibinfo{person}{Hugo Touvron}, {and} \bibinfo{person}{Herv{\'e} J{\'e}gou}.} \bibinfo{year}{2021}\natexlab{}.
\newblock \showarticletitle{Resnet strikes back: An improved training procedure in timm}.
\newblock \bibinfo{journal}{\emph{arXiv preprint arXiv:2110.00476}} (\bibinfo{year}{2021}).
\newblock


\bibitem[Wu et~al\mbox{.}(2021)]%
        {wu2021cvt}
\bibfield{author}{\bibinfo{person}{Haiping Wu}, \bibinfo{person}{Bin Xiao}, \bibinfo{person}{Noel Codella}, \bibinfo{person}{Mengchen Liu}, \bibinfo{person}{Xiyang Dai}, \bibinfo{person}{Lu Yuan}, {and} \bibinfo{person}{Lei Zhang}.} \bibinfo{year}{2021}\natexlab{}.
\newblock \showarticletitle{Cvt: Introducing convolutions to vision transformers}. In \bibinfo{booktitle}{\emph{ICCV}}. \bibinfo{pages}{22--31}.
\newblock


\bibitem[Xiao et~al\mbox{.}(2017)]%
        {xiao2017fashion-mnist}
\bibfield{author}{\bibinfo{person}{Han Xiao}, \bibinfo{person}{Kashif Rasul}, {and} \bibinfo{person}{Roland Vollgraf}.} \bibinfo{year}{2017}\natexlab{}.
\newblock \showarticletitle{Fashion-mnist: a novel image dataset for benchmarking machine learning algorithms}.
\newblock \bibinfo{journal}{\emph{arXiv preprint arXiv:1708.07747}} (\bibinfo{year}{2017}).
\newblock


\bibitem[Xie et~al\mbox{.}(2017)]%
        {xie2017dwconv}
\bibfield{author}{\bibinfo{person}{Saining Xie}, \bibinfo{person}{Ross Girshick}, \bibinfo{person}{Piotr Doll{\'a}r}, \bibinfo{person}{Zhuowen Tu}, {and} \bibinfo{person}{Kaiming He}.} \bibinfo{year}{2017}\natexlab{}.
\newblock \showarticletitle{Aggregated residual transformations for deep neural networks}. In \bibinfo{booktitle}{\emph{CVPR}}. \bibinfo{pages}{1492--1500}.
\newblock


\bibitem[Yu et~al\mbox{.}(2021)]%
        {yu2021s2mlpv2}
\bibfield{author}{\bibinfo{person}{Tan Yu}, \bibinfo{person}{Xu Li}, \bibinfo{person}{Yunfeng Cai}, \bibinfo{person}{Mingming Sun}, {and} \bibinfo{person}{Ping Li}.} \bibinfo{year}{2021}\natexlab{}.
\newblock \showarticletitle{S2-MLPv2: Improved Spatial-Shift MLP Architecture for Vision}.
\newblock \bibinfo{journal}{\emph{arXiv preprint arXiv:2108.01072}} (\bibinfo{year}{2021}).
\newblock


\bibitem[Yu et~al\mbox{.}(2022)]%
        {yu2022s2mlp}
\bibfield{author}{\bibinfo{person}{Tan Yu}, \bibinfo{person}{Xu Li}, \bibinfo{person}{Yunfeng Cai}, \bibinfo{person}{Mingming Sun}, {and} \bibinfo{person}{Ping Li}.} \bibinfo{year}{2022}\natexlab{}.
\newblock \showarticletitle{S2-mlp: Spatial-shift mlp architecture for vision}. In \bibinfo{booktitle}{\emph{Winter Conference on Applications of Computer Vision}}. \bibinfo{pages}{297--306}.
\newblock


\bibitem[Yun et~al\mbox{.}(2019)]%
        {yun2019cutmix}
\bibfield{author}{\bibinfo{person}{Sangdoo Yun}, \bibinfo{person}{Dongyoon Han}, \bibinfo{person}{Seong~Joon Oh}, \bibinfo{person}{Sanghyuk Chun}, \bibinfo{person}{Junsuk Choe}, {and} \bibinfo{person}{Youngjoon Yoo}.} \bibinfo{year}{2019}\natexlab{}.
\newblock \showarticletitle{Cutmix: Regularization strategy to train strong classifiers with localizable features}. In \bibinfo{booktitle}{\emph{ICCV}}. \bibinfo{pages}{6023--6032}.
\newblock


\bibitem[Zhang et~al\mbox{.}(2017)]%
        {zhang2017mixup}
\bibfield{author}{\bibinfo{person}{Hongyi Zhang}, \bibinfo{person}{Moustapha Cisse}, \bibinfo{person}{Yann~N Dauphin}, {and} \bibinfo{person}{David Lopez-Paz}.} \bibinfo{year}{2017}\natexlab{}.
\newblock \showarticletitle{mixup: Beyond empirical risk minimization}.
\newblock \bibinfo{journal}{\emph{arXiv preprint arXiv:1710.09412}} (\bibinfo{year}{2017}).
\newblock


\bibitem[Zhang et~al\mbox{.}(2018)]%
        {zhang2018shufflenet}
\bibfield{author}{\bibinfo{person}{Xiangyu Zhang}, \bibinfo{person}{Xinyu Zhou}, \bibinfo{person}{Mengxiao Lin}, {and} \bibinfo{person}{Jian Sun}.} \bibinfo{year}{2018}\natexlab{}.
\newblock \showarticletitle{Shufflenet: An extremely efficient convolutional neural network for mobile devices}. In \bibinfo{booktitle}{\emph{CVPR}}. \bibinfo{pages}{6848--6856}.
\newblock


\bibitem[Zhang et~al\mbox{.}(2022)]%
        {zhang2022nest}
\bibfield{author}{\bibinfo{person}{Zizhao Zhang}, \bibinfo{person}{Han Zhang}, \bibinfo{person}{Long Zhao}, \bibinfo{person}{Ting Chen}, \bibinfo{person}{Sercan~{\"O} Arik}, {and} \bibinfo{person}{Tomas Pfister}.} \bibinfo{year}{2022}\natexlab{}.
\newblock \showarticletitle{Nested hierarchical transformer: Towards accurate, data-efficient and interpretable visual understanding}. In \bibinfo{booktitle}{\emph{AAAI}}, Vol.~\bibinfo{volume}{36}. \bibinfo{pages}{3417--3425}.
\newblock


\bibitem[Zhong et~al\mbox{.}(2020)]%
        {zhong2020randerase}
\bibfield{author}{\bibinfo{person}{Zhun Zhong}, \bibinfo{person}{Liang Zheng}, \bibinfo{person}{Guoliang Kang}, \bibinfo{person}{Shaozi Li}, {and} \bibinfo{person}{Yi Yang}.} \bibinfo{year}{2020}\natexlab{}.
\newblock \showarticletitle{Random erasing data augmentation}. In \bibinfo{booktitle}{\emph{AAAI}}, Vol.~\bibinfo{volume}{34}. \bibinfo{pages}{13001--13008}.
\newblock


\end{thebibliography}

% \appendix

\clearpage
\begin{alphasection}
\section{Appendix}

\begin{figure*}[t]
  \centering
  \setcounter{figure}{0}
  \renewcommand{\thefigure}{A\arabic{figure}}
   %\includegraphics[width=0.8\linewidth]{egfigure.eps}
   \includegraphics[width=1.0 \linewidth]{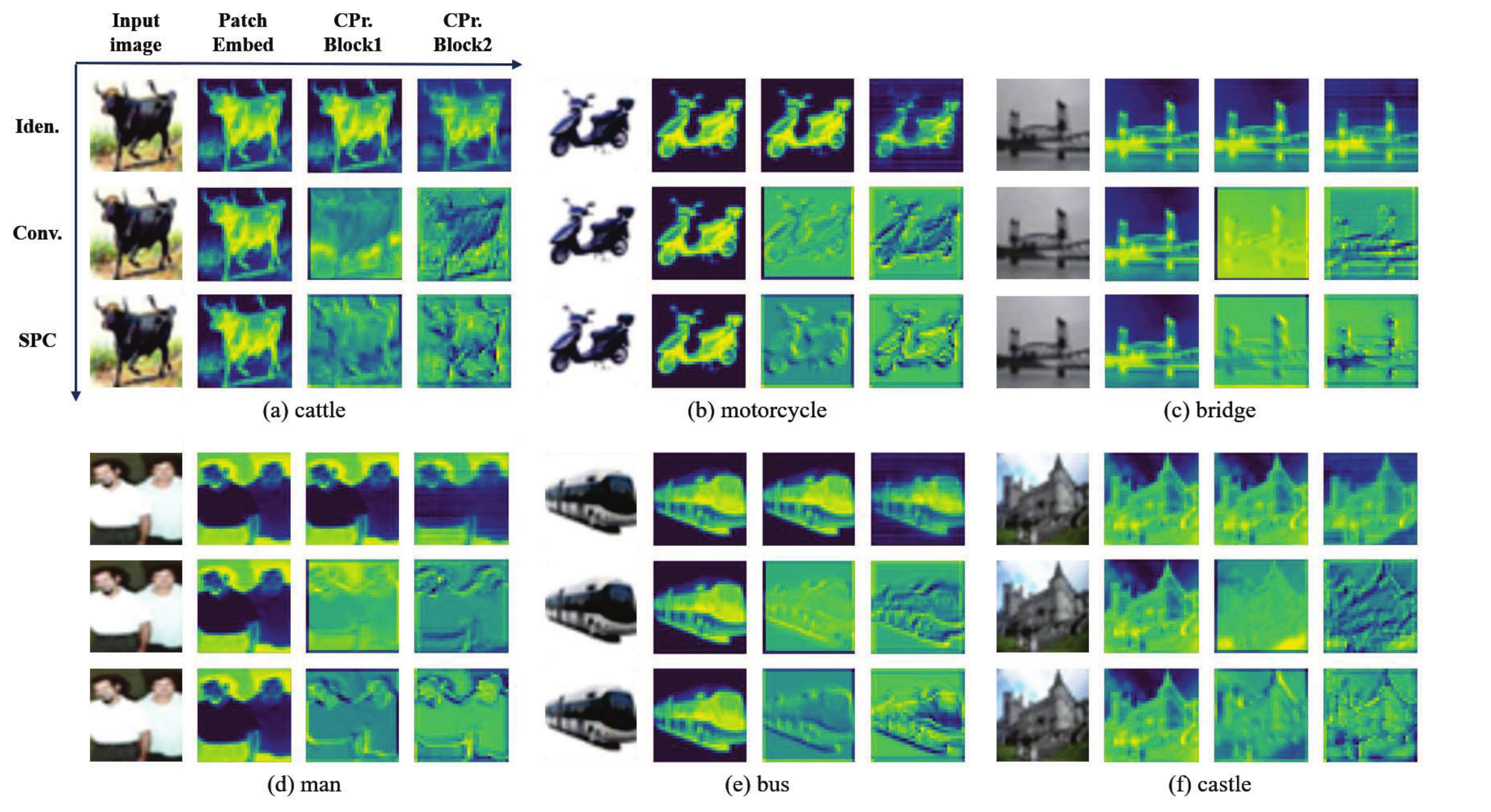}

   \caption{The feature maps of six samples with 3 rows and 4 columns. Each row represents a specific local modeling approach: identity (Iden.), convolution (Conv.) and SPC. The columns are the maps in different phases of Caterpillar (CPr.)-T.}
   \label{visualization}
\end{figure*}

\subsection{Supplementary Visualization}
\label{sup.1}
As referred to in Sec. 4.4, we illustrated the SPC, Conv and Iden. based Caterpillar in Fig. \ref{visualization}.

\subsection{Further Comparison with SOTA methods}
There are some representative MLP models and recent SOTA convolutional models have published.
We add a lot methods in Table \ref{tab-R1}. Due to the space limit, we only list the models with around 20-30M parameters.

    % \textcolor{red}{Please also show the results of the proposed method in Table R1}
\begin{table}[h]
  \setcounter{table}{0}
  \renewcommand{\thetable}{A\arabic{table}}
  \caption{Results (\%) on ImageNet-1K datasets.}
  \label{tab-R1}
  \centering
  \renewcommand\tabcolsep{10.0pt}
  \scalebox{0.9}{
  \begin{tabular}{l|c|c|c}
    \toprule
    Networks           & Params    & FLOPs     & Top-1   \\
    \midrule
    MLP-Mixer-B/16\cite{tolstikhin2021mlp-mixer}    & 30M       & -         & 76.4    \\
    CycleMLP-T\cite{chen2023cyclemlp}         & 28M       & 4.4G      & 81.3    \\
    ConvNeXt-T\cite{liu2022convnet}       & 29M       & 4.5G      & 82.1    \\
    SLak-T\cite{liumore}             & 30M       & 5.0G      & 82.5    \\
    VAN-B2\cite{guo2023visual}             & 27M       & 5.0G      & 82.8    \\
    HorNet-T\cite{rao2022hornet}           & 23M       & 3.9G      & 83.0    \\
    MogaNet-S\cite{li2023moganet}          & 25M       & 5.0G      & 83.4    \\
    InternImage-T\cite{wang2023internimage}      & 30M       & 5.0G      & 83.5    \\
   \rowcolor{blue!8} Caterpillar-T    & 29M    & 6.0G    & 82.4\\
    \bottomrule
  \end{tabular}
  }
\end{table}

We compare the models in three aspects:

\textbf{Performance.} The model performance are mainly determined by the architecture design and training strategies. 
For the recent SOTA methods like MogaNet, they mostly adopt richer strategies, such as larger batch size (i.e., 4096) and advanced augmentation (e.g., color jitter), which are effective but tricky. 
In our manuscript, we have compared Caterpillar with models with similar strategies, among which Caterpillar reached competitive or even superior performance. 
Therefore, despite the higher results of recent SOTA methods, Caterpillar is still competitive in classification capability.\\

\textbf{Technique.} Modern deep vision architectures can be decomposed into 5 levels, i.e., architecture, stage, block, module and layers, among which the spatial-mixing modules (in module level) mainly leads to the difference among various methods.
% The main difference among various models comes from the design of spatial modules (in module level), and more specifically, from the way to aggregate and combine local\&global information. 
MLP-Mixer and CycleMLP are early MLP models which focus on global and local information, respectively.
Another models are advanced conv-based models, which all make full use of depth-wise convolution (DWConv), with both local and global information aggregated in parallel regimes.
Caterpillar separately captures the local and global information through SPC and sparse-MLP modules. \\

\textbf{Trend.} Recent convolutional and MLP models have shown similar trends, i.e., capturing spatial information more elaborately and sparsely, from MLP-mixer, through ResMLP and to sMLPNet in MLP family, and from Conv-Mixer, through ConvNeXt, and to Moganet in Conv models. Therefore, combining lightweight techniques like depth-wise with SPC module could be promising works.

\subsection{Further Explanation for Shift Methods}
\textit{Shift} spawns a broad class of methods, where the shifting steps, directions, dimensions as well as the restoring/learning ways mainly determines the various forms and functions for those methods.
We add a detailed comparison of the SPC with some shift-base methods of different operations and applications.

\textbf{Comparison with AS-MLP.} 
The main difference between AS-MLP\cite{lian2021asmlp} and Caterpillar is shifting dimensions and directions. AS-MLP shifts each channel of the image along two directions, while Caterpillar shifts the entire image into four neighboring maps and encode them in individual sub-spaces, achieving parallel computing capability.

% \textcolor{red}{Too long. It should directly show the difference of SPC to GSTS and X-volution.}
\textbf{Comparison with Grouped Spatial-temporal Shift (GSTS) and X-volution.}
Both of the shifting operations in GSTS\cite{li2023simple} and X-volution\cite{chen2021x} are adopted to aggregate long-term information, with the GSTS shifted in both temporal and spatial dimensions for 3D data and X-volution shifted in more flexible directions (e.g., left-righ) for wider receptive field.
% The shift in GSTS is for 3D data with longer shifting steps in both temporal and spatial dimensions, while the operation in X-volution is conducted with more flexible shifting directions (e.g., left-righ) that can obtain wider receptive field.
% Additionally, both the shift operations in GSTS and X-volution are combined with extra operations (e.g., convolution, self-attention) to aggregate long-term information. 
Different from them, the SPC is proposed with a 4-scoped receptive field to capture local information elaborately.

\subsection{Efficiency Concerns}
\textbf{Further comparison between Caterpillar, sMLPNet and Strip-MLP.}
Both Strip-MLP\cite{cao2023stripmlp} and Caterpillar are built upon sMLPNet\cite{tang2022sMLPNet}. 
Strip-MLP replaces the sMLP with Strip-MLP layer in sMLPNet and lightweights the model architectures, 
while Caterpillar only replaces the DWConv with SPC module and thus has more parameters.
However, the Caterpillar-T$^{\dag}$ (\textit{e.g.,} 23M, 77.6\% on MIN) with similar computational costs can also obtain higher results to Strip-MLP-T (23M, 76.5\%) and sMLPNet-T (24M, 77.1\%) (Table \ref{tab-result-M1}), implying the possibility of Caterpillar to be lightweight versions.

\textbf{More evaluation about efficiency.}
We add the comparison of SPC and conv-based ResNet in Table \ref{tab-R2}.
As we can see, SPC module brings lower computational complexity than convolution, since SPC-based model achieves lower parameters and FLOPs. 
SPC module also provides higher Thourghput and less training times.
% As we can see, SPC module brings lower parameters and FLOPs, as well as higher Thourghput and less training times than convolution, showing that SPC achieves lower computational complexity than convolution
However, such trend is not with directly proportional relationship, e.g., 1/4 parameters is not bringing 4 times boosting of Throughput or 1/4 training time. 
This could be attribute to the lower computation \textit{vs.} memory access rate in SPC module, i.e., memory access may takes more execution time than computation can cannot fully utilize the GPU capacity——the similar problem in depthwise separable convolution\cite{zhang2018shufflenet,qin2018diagonalwise}.
% the linear and concatenation adopted in SPC module, as such operations always limited in hardware computation.
We hope further optimizations on hardware computation can further improve its computational efficiency.

\begin{table}[h]
  \setcounter{table}{1}
  \renewcommand{\thetable}{A\arabic{table}}
  \caption{Comparison between SPC- and Conv-based ResNet.}
  \label{tab-R2}
  \centering
  \renewcommand\tabcolsep{8.0pt}
  \scalebox{0.85}{
  \begin{tabular}{l|c|c|c|c}
    \toprule
    Networks           & Params    & FLOPs   & Throughput   & Training Time   \\
    \midrule
    Res-18              & 12M      & 1.8G    & 2174         &  20 hours   \\
    Res-18 (SPC)         & 3M      & 0.6G    & 4885         &  14 hours   \\
    \bottomrule
  \end{tabular}
  }
\end{table}

\subsection{Detailed Architecture Specifications}
\label{sup:detailArch}

As mentioned in Section \ref{Section 3.3}, we build our tiny, small, and base models called Caterpillar-T, -S, -B, which adopt identical backbone architectures to {sMLPNet-T}, -S, -B, respectively. The only difference between the Caterpillar and sMLPNet is the local-mixing ways (\textit{i.e.}, SPC vs DWConv). To enable Caterpillar more friendly to limited computational resources, we also introduce the mini and tiny\_x models of Caterpillar, namely Caterpillar-Mi and -Tx, which are variants of about 0.2 $\times$ , 0.5 $\times$ the parameters and FLOPs of the -T model. Table \ref{tab-cprArchitectures} displays their detailed architectures.

\begin{table*}[h]
  \renewcommand{\thetable}{A\arabic{table}}
  \caption{Detailed settings of Caterpillar series.}
  \label{tab-cprArchitectures}
  \centering
  \renewcommand\tabcolsep{3.0pt}
  % \resizebox{\textwidth}{!}{
  \renewcommand\arraystretch{1.2}
  \begin{tabular}{c|c|c|c|c|c}
    \toprule
    Stages               & Caterpillar-Mi   & Caterpillar-Tx   & Caterpillar-T   & Caterpillar-S    & Caterpillar-B  \\
    \midrule
    \multirow{2}*{S1}    & patch\_size: 4    & patch\_size: 4   & patch\_size: 4  & patch\_size: 4   & patch\_size: 4      \\
    ~                    & [56×56, 40]×2     & [56×56, 60]×2     & [56×56, 80]×2    & [56×56, 96]×2     & [56×56, 112]×2     \\
    \hline
    \multirow{2}*{S2}    & downsp. rate: 2   & downsp. rate: 2   & downsp. rate: 2  & downsp. rate: 2   & downsp. rate: 2     \\
    ~                    & [28×28, 80]×6     & [28×28, 120]×8    & [28×28, 160]×8   & [28×28, 192]×10   & [28×28, 224]×10     \\
    \hline
    \multirow{2}*{S3}    & downsp. rate: 2   & downsp. rate: 2   & downsp. rate: 2  & downsp. rate: 2   & downsp. rate: 2     \\
    ~                    & [14×14, 160]×10   & [14×14, 240]×14   & [14×14, 320]×14  & [14×14, 384]×24   & [14×14, 448]×24     \\
    \hline
    \multirow{2}*{S4}    & downsp. rate: 2   & downsp. rate: 2   & downsp. rate: 2  & downsp. rate: 2   & downsp. rate: 2     \\
    ~                    & [7×7, 320]×2      & [7×7, 480]×2      & [7×7, 640]×2     & [7×7, 768]×2      & [7×7, 896]×2     \\
    \bottomrule
  \end{tabular}
  % }
\end{table*}

\begin{table*}[b]
  \renewcommand{\thetable}{A\arabic{table}}
  \caption{Feature maps in models with different architectures on four small-scale benchmarks. C denotes the channel number of the used models in their first stage.}
  \label{tab-smallMap}
  \centering
  \renewcommand\arraystretch{1.0}
  \begin{tabular}{l|c|cccc}
    \toprule
    Architecture  & Stages   & MIN   & C10   & C100    & Fashion  \\
    \midrule
    \multirow{4}*{Pyramid}  & S1       & [28×28, C]        & [32×32, C]       & [32×32, C]        & [28×28, C]    \\
    ~             & S2       & [14×14, 2C]       & [16×16, 2C]      & [16×16, 2C]       & [14×14, 2C]   \\
    ~             & S3       & [7×7, 4C]         & [8×8, 4C]        & [8×8, 4C]         & [7×7, 4C]     \\
    ~             & S4       & [7×7, 8C]         & [4×4, 8C]        & [4×4, 8C]         & [7×7, 8C]     \\
    \midrule
    \multirow{2}*{2Stage}        & S1       & [14×14, C]       & [16×16, C]      & [16×16, C]       & [14×14, C]   \\
    ~             & S2       & [7×7, 2C]         & [8×8, 2C]        & [8×8, 2C]         & [7×7, 2C]     \\
    \midrule
    Isotropic     & S1       & [7×7, C]         & [8×8, C]        & [8×8, C]         & [7×7, C]     \\
    \bottomrule
  \end{tabular}
\end{table*}

\subsection{Implementation of Models on Small Images}
\label{sup:Imp_SmallImage}

In Section \ref{Section 4.1}, we have conducted fifteen vision models on small-scale image recognition tasks.
Among them, \cite{touvron2022resmlp,trockman2022convmixer,touvron2021deit,hassani2021cct} are built with isotropic structure, \cite{hou2022vip,wang2022dynamixer} are with 2Stage structure,  \cite{chen2023cyclemlp,guo2022hiremlp,tang2022wavemlp,he2016resnet,liu2021swin,zhang2022nest,tang2022sMLPNet,cao2023stripmlp} and Caterpillar are with pyramid structure.
For fair comparison (\textit{i.e.,} enabling the parameters and FLOPs of these models to be similar), we set the \textit{patch size} to 3, 1, 1, 1 in their \textit{patch embedding layer} for pyramid models when applied on the MIN, C10, C100 and Fashion datasets, 6, 2, 2, 2 for 2Stage models, and 12, 4, 4, 4 for Isotropic models, respectively. The feature maps in their main computational bodies on the four datasets are listed in Table \ref{tab-smallMap}.

\subsection{Training Strategies}
\label{sup:trainingStrategies}

In Table \ref{tab-trainSche}, we present the training strategies for all models adopted in Section \ref{Section 4.1}. 
These strategies are the same as those in their original papers for ImageNet-1k training. 
Note that we don’t employ ‘EMA’ for small-scale image recognition studies, since it decreases the performance of all models by a large margin. 
For the proposed Caterpillar, we list its training procedure in Table \ref{tab-trainScheCPr}, which is applied for both ImageNet-1K and small-scale benchmarks.
% For the ImageNet-1k classification task (Section \ref{Section 4.3}), we set the ‘EMA’ to 0.99996.

\begin{table*}[hb]
  \renewcommand{\thetable}{A\arabic{table}}
  \caption{Training strategies for Caterpillar models}
  \label{tab-trainScheCPr}
  \centering
  \renewcommand\tabcolsep{12.0pt}
  \renewcommand\arraystretch{1.0}
  % \resizebox{\textwidth}{!}{
  \begin{tabular}{l|c}
    \toprule
    \toprule
    \multirow{2}*{Configs}   & Caterpillar \\
    ~     & Mi, Tx, T, S, B \\
    \midrule
    Training epochs & 300  \\
    Batch size      & 1024 \\
    Optimizer       & AdamW \\
    LR              & 1e-3 \\
    LR decay        & cosine \\
    Min LR          & 1e-5 \\
    Weight\_decay   & 0.05 \\
    Warmup epochs   & 5 \\
    Warmup LR       & 1e-6\\
    \midrule
    Rand Augment    & 9/0.5\\
    Mixup           & 0.8\\
    Cutmix          & 1.0\\
    Stoch. Depth    & 0, 0, 0.05, 0.2, 0.3\\
    Repeated Aug    & \Checkmark\\
    Erasing prob.   & 0.25 \\
    Label smoothing & 0.1 \\
    \textit{EMA}             & \textit{0.99996} \\
    \bottomrule
    \bottomrule
  \end{tabular}
  % }
\end{table*}

\begin{table*}[t]
  \renewcommand{\thetable}{A\arabic{table}}
  \caption{Training strategies for various vision models}
  \label{tab-trainSche}
  \centering
  \renewcommand\tabcolsep{3.0pt}
  \renewcommand\arraystretch{1.0}
  \resizebox{\textwidth}{!}{
  \begin{tabular}{l|ccccccc}
    \toprule
    \toprule
    \multirow{2}*{Configs}  & ResNet   & ConvMixer   & DeiT   & Swin    & CCT  & NesT    & ResMLP \\
    
    ~             & 18, 34, 50 \cite{wightman2021resnetstrike}       & 768/32 \cite{trockman2022convmixer}      & T, S \cite{touvron2021deit}      & T \cite{liu2021swin}     & 7/3$\times$1 \cite{hassani2021cct}   & T \cite{zhang2022nest}       &S12, S24 \cite{touvron2022resmlp} \\
    \midrule
    Training epochs        & 300       & 300      & 300      & 300       & 300      & 300       & 400  \\
    Batch size             & 2048      & 640      & 1024     & 1024      & 1024     & 512       & 1024 \\
    Optimizer              & LAMB      & AdamW    & AdamW    & AdamW     & AdamW    & AdamW     & LAMB \\
    LR                     & 5e-3      & 1e-2     & 1e-3     & 1e-3      & 5e-4     & 5e-4      & 5e-3 \\
    LR decay               & cosine    & onecycle   & cosine   & cosine    & cosine   & cosine    & cosine \\
    Min LR                 & 1e-6      & 1e-6     & 1e-5     & 5e-6      & 1e-5     & 0         & 1e-5 \\
    Weight\_decay          & 0.02      & 0.00002  & 0.05     & 0.05      & 0.05     & 0.05      & 0.2 \\
    Warmup epochs          & 5         & 0        & 5        & 20        & 10       & 20        & 5 \\
    Warmup LR              & 1e-4      & --        & 1e-6     & 5e-7      & 1e-6     & 1e-6      & 1e-6\\
    \midrule
    Rand Augment           & 7/0.5     & 9/0.5    & 9/0.5    & 9/0.5     & 9/0.5    & 9/0.5     & 9/0.5\\
    Mixup                  & 0.1       & 0.5      & 0.8      & 0.8       & 0.8      & 0.8       & 0.8\\
    Cutmix                 & 1.0       & 0.5      & 1.0      & 1.0       & 1.0      & 1.0       & 1.0\\
    Stoch. Depth           & 0.05      & 0        & 0.1      & 0.2       & 0        & 0.2       & 0.1\\
    Repeated Aug           & \Checkmark  &  \textcolor{lightgray}{\XSolidBrush}        & \Checkmark  &  \textcolor{lightgray}{\XSolidBrush}         &  \textcolor{lightgray}{\XSolidBrush}        &  \textcolor{lightgray}{\XSolidBrush}         & \Checkmark\\
    Erasing prob.          & 0         & 0.25     & 0.25     & 0.25      & 0.25     & 0.25      & 0.25 \\
    Label smoothing        & 0         & 0.1      & 0.1      & 0.1       & 0.1      & 0.1       & 0.1 \\
    \textit{EMA}           & --       & --          & --  & --         & --        & --         & --\\
    \bottomrule

    \bottomrule
    \multirow{2}*{Configs}               & CycleMLP  & HireMLP  & Wave-MLP  & ViP     & DynaMixer  & sMLPNet   & Strip-MLP \\
    
    ~     & B1, B2 \cite{chen2023cyclemlp}    & Ti, S \cite{guo2022hiremlp}    & T, S \cite{tang2022wavemlp}     & S7 \cite{hou2022vip}      & S \cite{wang2022dynamixer}          & T \cite{tang2022sMLPNet} & T*, T \cite{cao2023stripmlp}\\
    \midrule
    Training epochs & 300       & 300        & 300      & 300   & 300        & 300       & 300  \\
    Batch size      & 1024      & 2048, 1024 & 1024     & 2048  & 2048       & 1024      & 1024 \\
    Optimizer       & AdamW     & AdamW      & AdamW    & AdamW & AdamW      & AdamW     & AdamW \\
    LR              & 1e-3      & 1e-3     & 1e-3     & 2e-3      & 2e-3     & 1e-3      & 1e-3 \\
    LR decay        & cosine    & cosine   & cosine   & cosine    & cosine   & cosine    & cosine \\
    Min LR          & 1e-5      & 1e-5     & 1e-5     & 1e-5      & 1e-5     & 1e-5      & 5e-6 \\
    Weight\_decay   & 0.05      & 0.05     & 0.05     & 0.05      & 0.05     & 0.05      & 0.05 \\
    Warmup epochs   & 5         & 20       & 5        & 20        & 20       & 20        & 30 \\
    Warmup LR       & 1e-6      & 1e-6     & 1e-6     & 1e-6      & 1e-6     & 1e-6      & 5e-7\\
    \midrule
    Rand Augment    & 9/0.5     & 9/0.5    & 9/0.5    & 9/0.5     & 9/0.5    & 9/0.5     & 9/0.5\\
    Mixup           & 0.8      & 0.8       & 0.8      & 0.8       & 0.8      & 0.8       & 0.8\\
    Cutmix          & 1.0       & 1.0      & 1.0      & 1.0       & 1.0      & 1.0       & 1.0\\
    Stoch. Depth    & 0.1      & 0         & 0.1      & 0.1       & 0.1      & 0         & 0.2\\
    Repeated Aug    & \Checkmark  & \Checkmark & \Checkmark & \textcolor{lightgray}{\XSolidBrush}  &  \textcolor{lightgray}{\XSolidBrush}  & \Checkmark  & \textcolor{lightgray}{\XSolidBrush}\\
    Erasing prob.   & 0.25      & 0.25     & 0.25     & 0.25      & 0.25     & 0.25      & 0.25 \\
    Label smoothing & 0.1       & 0.1      & 0.1      & 0.1       & 0.1      & 0.1       & 0.1 \\
    \textit{EMA}             & \textit{0.99996}   & --        & \textit{0.99996}  & --         & \textit{0.99996}  & \textit{0.99996}   &  --\\
    \bottomrule
    \bottomrule
  \end{tabular}
  }
\end{table*}

\subsection{Sparse-MLP Module}
\label{sup:sMLP_module}

The sparse-MLP (sMLP) module is proposed in \cite{tang2022sMLPNet} and also adopted in the Caterpillar block for aggregating global information. To have a comprehensive understanding of the proposed Caterpillar, we also depict the sMLP module in Figure \ref{fig-smlp}. 
As we can see, the sMLP module consists of three branches: two of them are used to mix information along horizontal and vertical directions, respectively, which is implemented by two H (W) $\times$ H (W) linear projections, and the other path is an identity mapping. The output of the three branches are concatenated and then mixed by a 3C $\times$ C linear projection to obtain the final output.
Through the sMLP calculation, each pillar can gather information from other pillars in the same row and column.
Stacking more sMLP blocks allows for the mixing of the gathered features across different rows and columns, with all pillars incorporating the global information of the entire image.

\begin{figure*}[h]
  \begin{minipage}[c]{\textwidth}
  \centering
    \renewcommand{\thefigure}{A\arabic{figure}}
    \includegraphics[width=0.95 \linewidth]{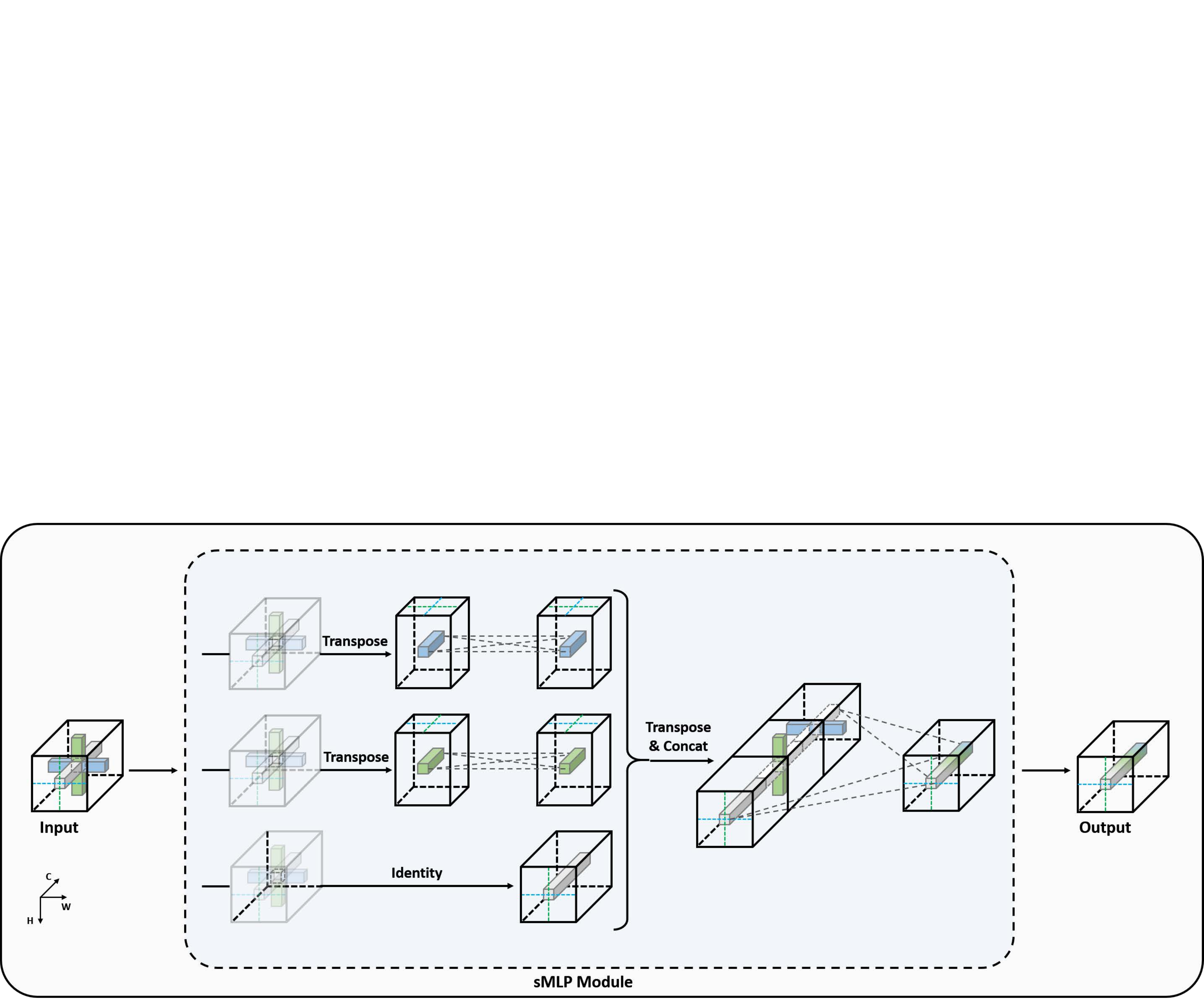}
    \vspace*{\fill}
    \caption{The sparse-MLP module proposed in sMLPNet \cite{tang2022sMLPNet}}
   \label{fig-smlp}
   \end{minipage}
\end{figure*}

% Specifically, global information can be aggregated for each pillar through at least two executions (layers) of the sMLP calculation. 
% In the first execution, each pillar collects information from other pillars in the same row and column. 
% In the next execution, the gathered features from corresponding rows and columns are mixed again. 
% Stacking this process enables integration across different rows and columns, with all pillars incorporated of global information for the entire image.
% This process ensures the incorporation of global information for the entire image on all pillars.
% For each pillar, global information is aggregated through two executions of the sMLP calculation:
% In the first execution, each pillar gathers information from other ones in the same row and column.
% In the second execution, the pillars, which have integrated corresponding row and column features, are mixed again, allowing for the integration of features across different rows and columns and thus incorporating global information for the entire image on all pillars.
% the sMLP calculation allows for the integration of features across different rows and columns, with global information of the entire image incorporated into each pillar.

\end{alphasection}

\end{document}